\documentclass[final]{cvpr}

\usepackage{times}
\usepackage{epsfig}
\usepackage{graphicx}
\usepackage{amsmath}
\usepackage{amssymb}
\usepackage{multirow}
\usepackage{ulem}
\usepackage{arydshln}
\usepackage{bm}
\usepackage{booktabs}
\usepackage[table]{xcolor}
\usepackage{makecell}
\usepackage{diagbox}
\usepackage[font=small,labelfont=bf]{caption}

\usepackage[pagebackref=true,breaklinks=true,letterpaper=true,colorlinks,bookmarks=false]{hyperref}

\newcommand{\CUT}[1]

\def\cvprPaperID{724} 
\def\confYear{CVPR 2021}

\begin{document}

\title{Model-based 3D Hand Reconstruction via Self-Supervised Learning}

\author{
{Yujin Chen$^{1}$\thanks{Work done during an internship at Tencent AI Lab.}\qquad Zhigang Tu$^{1}$\thanks{Corresponding author: tuzhigang@whu.edu.cn}\qquad  Di Kang$^{2}$\qquad  Linchao Bao$^{2}$}\\ 
{Ying Zhang$^{3}$\quad \,  Xuefei Zhe$^{2}$\quad \,  Ruizhi Chen$^{1}$\quad\,  Junsong Yuan$^{4}$}\\
	$^{1}$Wuhan University\quad
	$^{2}$Tencent AI Lab\quad
	$^{3}$Tencent\quad
	$^{4}$State University of New York at Buffalo\\
	{\tt\small \{yujin.chen, tuzhigang, ruizhi.chen\}@whu.edu.cn \{di.kang, zhexuefei\}@outlook.com}\\
	{\tt\small  linchaobao@gmail.com yinggzhang@tencent.com  jsyuan@buffalo.edu}
}
\maketitle

\begin{abstract}
Reconstructing a 3D hand from a single-view RGB image is challenging due to various hand configurations and depth ambiguity.
To reliably reconstruct a 3D hand from a monocular image, most state-of-the-art methods heavily rely on 3D annotations at the training stage, but obtaining 3D annotations is expensive.
To alleviate reliance on labeled training data, we propose ${\rm {S}^{2}HAND}$, a self-supervised 3D hand reconstruction network that can jointly estimate pose, shape, texture, and the camera viewpoint.
Specifically, we obtain geometric cues from the input image through easily accessible 2D detected keypoints.
To learn an accurate hand reconstruction model from these noisy geometric cues, we utilize the consistency between 2D and 3D representations and propose a set of novel losses to rationalize outputs of the neural network.
For the first time, we demonstrate the feasibility of training an accurate 3D hand reconstruction network without relying on manual annotations.
Our experiments show that the proposed self-supervised method achieves comparable performance with recent fully-supervised methods.
The code is available at {\url{https://github.com/TerenceCYJ/S2HAND}}.
\vspace{-0.1in}
\end{abstract}
\begin{figure}[th]
\vspace{-0.2cm}
\begin{center}
\includegraphics[width=1\linewidth]{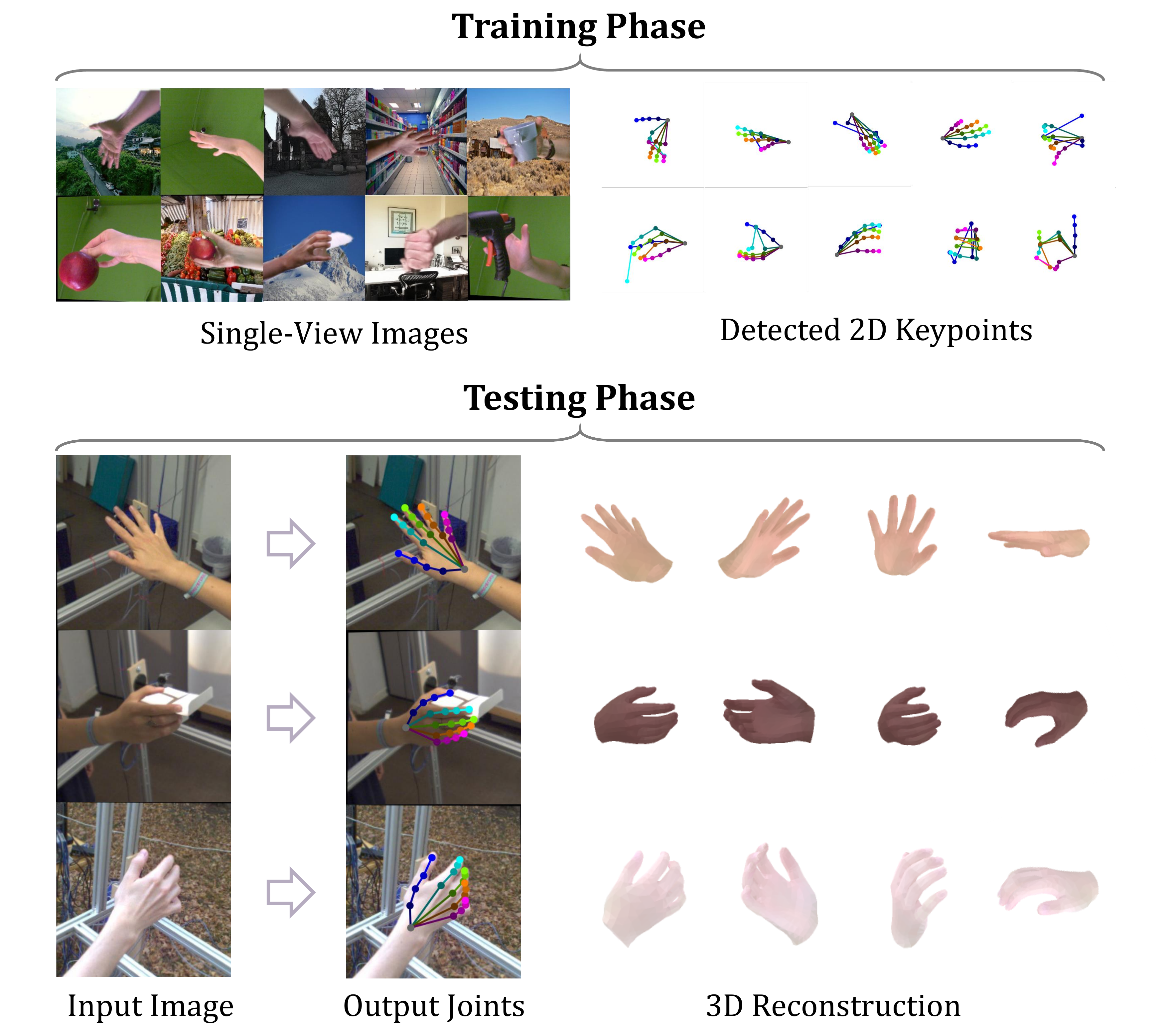}
\captionof{figure}{Given a collection of unlabeled hand images, we learn a 3D hand reconstruction network in a self-supervised manner.
Top: the training uses a collection of unlabeled hand images and their corresponding noisy detected 2D keypoints.
Bottom: our model outputs accurate hand joints and shapes, as well as vivid textures.}
\label{fig:1}
\end{center}
\vspace{-0.8cm}
\end{figure}

\section{Introduction}
%
\begin{table}[tb]
\setlength{\abovecaptionskip}{0.cm}
\centering
\scalebox{0.85}{
\begin{tabular}{cll}
\Xhline{1pt}
\rowcolor[rgb]{ .873,  .91,  0.95}
Approach & Supervision & Outputs\\
\hline
\cite{qian2020parametric}& 3DM, 3DJ, 2DKP, I, TI & 3DM, 3DJ, Tex\\
\rowcolor[rgb]{ .94,  .94,  .94}
\cite{hasson2019learning,moon2020i2l}& 3DM, 3DJ & 3DM, 3DJ\\
\cite{yang2020seqhand}& 3DM*, 3DJ, 2DKP, 2DS, Syn& 3DM, 3DJ\\
\rowcolor[rgb]{ .94,  .94,  .94}
\cite{Ge_2019_CVPR}& 3DM*, 3DJ, 2DKP, D*& 3DM, 3DJ\\
\cite{Kulon_2020_CVPR}& 3DM*, D2DKP & 3DM, 3DJ\\
\rowcolor[rgb]{ .94,  .94,  .94}
\cite{zhou2020monocular}& 3DJ, 2DKP, Mo & 3DM, 3DJ\\
\cite{Baek_2019_CVPR,boukhayma20193d,Zhang_2019_ICCV}& 3DJ, 2DKP, 2DS & 3DM, 3DJ \\
\rowcolor[rgb]{ .94,  .94,  .94}
\cite{zimmermann2017learning}& 3DJ, 2DKP, 2DS & 3DJ \\
\cite{iqbal2018hand}& 3DJ, 2DKP & 3DJ\\
\rowcolor[rgb]{ .94,  .94,  .94}
\cite{Baek_2020_CVPR}& 3DJ*, 2DKP, 2DS & 3DM, 3DJ, Tex\\
\cite{spurr2020weakly}& 3DJ*, 2DKP & 3DJ\\
\rowcolor[rgb]{ .94,  .94,  .94}
\cite{cai2018weakly}& 2DKP, D & 3DJ\\
Ours & D2DKP, I & 3DM, 3DJ, Tex\\
\Xhline{1pt}
\end{tabular}}
\vspace{0.2cm}
\caption{A comparison of some representative 3D hand recovery approaches with highlighting the differences between the supervision and the outputs. We use the weakest degree of supervision and output the most representations. 3DM: 3D mesh, 3DJ: 3D joints, I: input image, TI: an additional set of images with clear hand texture, Tex: texture, 2DKP: 2D keypoints, 2DS: 2D silhouette, D: depth, D2DKP: detected 2D keypoints, Syn: extra synthetic sequence data, Mo: extra motion capture data. 
* indicates that the study uses multiple datasets for training, and at least one dataset used the supervision item.
}
\label{table:supervision}
\vspace{-0.4cm}
\end{table}
Reconstructing 3D human hands from a single image is important for computer vision tasks such as hand-related action recognition, augmented reality, sign language translation, and human-computer interaction \cite{holl2018efficient,parelli2020exploiting,tu2019action}. 
However, due to the diversity of hands and the depth ambiguity in monocular 3D reconstruction, image-based 3D hand reconstruction remains a challenging problem.

In recent years, we have witnessed fast progress in recovering 3D representations of human hands from images.
In this field, most methods were proposed to predict 3D hand pose from the depth image \cite{armagan2020measuring,chen2019so,ge2016robust,huang2020hand,yuan2018depth} or the RGB image \cite{athitsos2003estimating, cai2018weakly,iqbal2018hand,spurr2020weakly,zimmermann2017learning}. 
However, the surface information is needed in some applications such as grasping an object by a virtual hand \cite{holl2018efficient}, where the 3D hand pose represented by sparse joints is not sufficient.
To better display the surface information of the hand, previous studies predict the triangle mesh either via regressing per-vertex coordinate \cite{Ge_2019_CVPR,Kulon_2020_CVPR} or by deforming a parametric hand model \cite{hasson2020leveraging,hasson2019learning}.
Outputting such high-dimensional representation from 2D input is challenging for neural networks to learn, thus resulting in the training process relying heavily on 3D annotations such as dense hand scans, model-fitted parametric hand mesh, or human-annotated 3D joints.
Besides, the hand texture is important in some applications, such as vivid hands reconstruction in immersive virtual reality.
But only recently has a study exploring parametric texture estimation in a learning-based hand recovery system \cite{qian2020parametric}, while most previous works do not consider texture modeling.

Our key observation is that the 2D cues in the image space are closely related to the 3D hand model in the real world.
The 2D hand keypoints contain rich structural information, and the image contains texture information.
Both are important for reducing the use of expensive 3D annotations but have not been investigated much.
In this way, we could directly use 2D annotations and the input image to learn the structural and texture representations without using 3D annotations.
However, it is still labor-consuming to annotate 2D hand keypoints.
To completely save the cost of manual annotation, we propose to extract some geometric representations from the unlabeled image to help shape reconstruction and use the texture information contained in the input image to help texture modeling.

Motivated by the above observations, this work seeks to train an accurate and robust 3D hand reconstruction network only using supervision signals obtained from the input images and eliminate all manual annotations, which is the first attempt in this task.
To this end, we use an off-the-shelf 2D keypoint detector \cite{cao2019openpose} to produce some noisy 2D keypoints and supervise the hand reconstruction by these noisy detected 2D keypoints and the input image.
To better achieve this goal, there are several issues that need to be addressed. 
First, how to efficiently use joint-wise 2D keypoints to supervise the ill-posed monocular 3D hand reconstruction?
Second, since our setting does not use any ground truth annotation, how do we handle the noise in the 2D detection output?
%

To address the first issue, a model-based autoencoder is presented to estimate 3D joints and shape, where the output 3D joints are projected into image space and forced to align with the detected keypoints during training.
However, if we only align keypoints in image space, invalid hand poses often occur. This may be an invalid 3D hand configure that could be projected to be the correct 2D keypoints.
Also, 2D keypoints cannot reduce the scale ambiguity of the predicted 3D hand.
Thus, we design a series of priors embedded in the model-based hand representations to help the neural network output hand with a reasonable pose and size.
To address the second issue, a trainable 2D keypoint estimator and a novel 2D-3D consistency loss are proposed. The 2D keypoint estimator outputs joint-wise 2D keypoints and the 2D-3D consistency loss links the 2D keypoint estimator and the 3D reconstruction network to make the two mutually beneficial to each other during the training.
In addition, we find that the detection accuracy of different samples varies greatly, thus we propose to distinguish each detection item to weigh its supervision strength accordingly.

In summary, we present a $\mathbf{S^2HAND}$ (self-supervised 3D hand reconstruction) model which enables us to train a neural network that can predict 3D pose, shape, texture, and camera viewpoint from a hand image without any ground truth annotation, except that we use the outputs from a 2D keypoint detector (Fig.~\ref{fig:1}). 

Our main contributions are summarized as follows:
\vspace{-0.3cm}
\begin{itemize}
\setlength{\itemsep}{0pt}
\setlength{\parsep}{0pt}
\setlength{\parskip}{0pt}
\setlength{\topsep}{0pt}
\setlength{\partopsep}{0pt}
\item [$\bullet$] We present the first self-supervised 3D hand reconstruction network, which accurately outputs 3D joints, mesh, and texture from a single image, without using any annotated training data.
\item [$\bullet$] We exploit an additional trainable 2D keypoint estimator to boost the 3D reconstruction through a mutual improvement manner, in which a novel 2D-3D consistency loss is proposed.
\item [$\bullet$] We introduce a hand texture estimation module to learn vivid hand texture through self-supervision.
\item [$\bullet$] We benchmark self-supervised 3D hand reconstruction on some currently challenging datasets, where our self-supervised method achieves comparable performance to previous fully-supervised methods.

\end{itemize}
\vspace{-0.3cm}
%
\begin{figure*}[t]
\vspace{-0.6cm}
    \setlength{\abovecaptionskip}{0.cm}
	\begin{center}
		\includegraphics[width=0.98\linewidth]{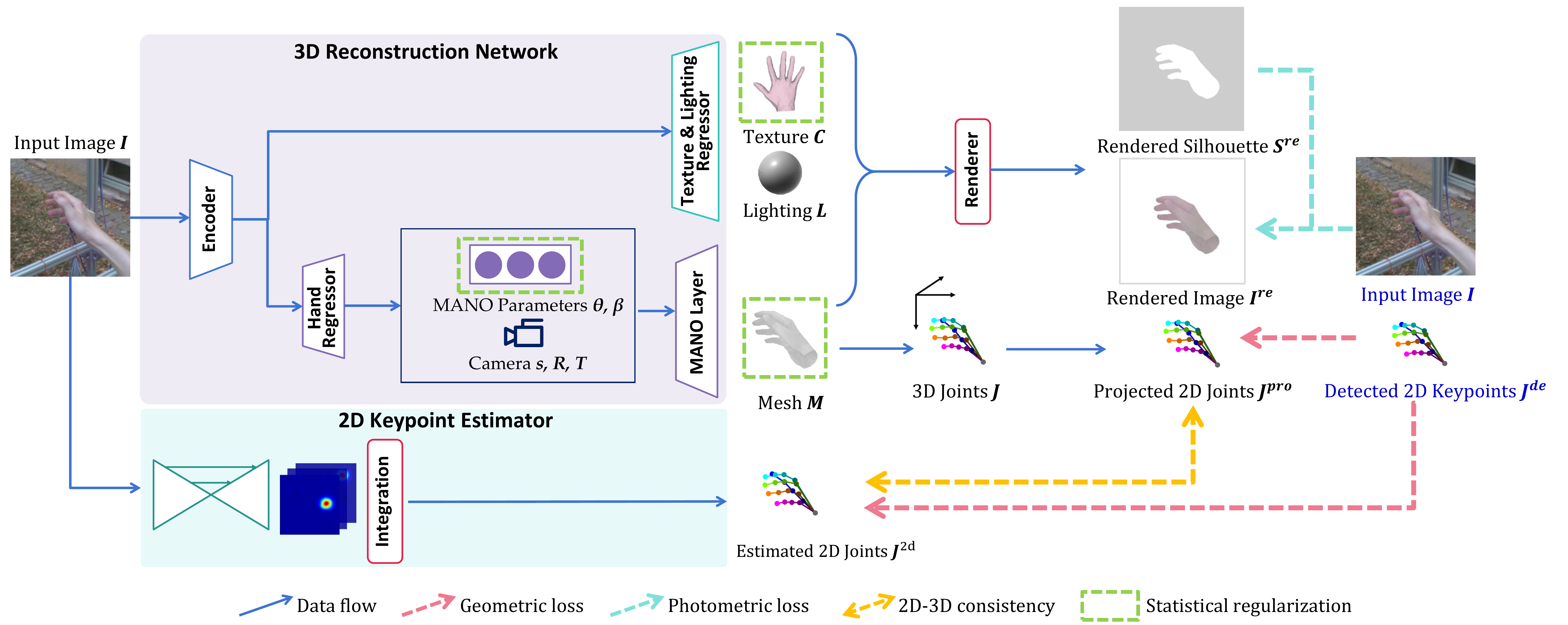}
	\end{center}
	\vspace{-0.2cm}
	\caption{Overview of the proposed framework.
	Our 3D reconstruction network decomposes an input image $I$ into pose, shape, viewpoint, texture, and lighting. The network is trained to reconstruct the input hand image and align with the detected 2D keypoints without extra ground truth annotation. We also adopt an additional trainable 2D keypoint estimator for joint-wise 2D keypoint estimation, which is supervised by the detected 2D keypoints as well. If the 2D keypoint estimator is enabled, a 2D-3D consistency function is introduced to link the 2D and 3D components for mutual improvement.
	During the inference, only the 3D reconstruction network is utilized.
	The ``Input Image $I$" and \mbox{``Detected 2D Keypoints $J^{de}$"} on the right side of this figure are used to calculate losses.
	}
	\label{fig:pipeline}
\vspace{-0.3cm}
\end{figure*}

\section{Related Work}
In this section, we review previous works that are related to our approach.
Our focus is on model-based 3D hand pose and shape estimation, and 3D reconstruction with limited supervision.
For more work on 3D pose estimation, please refer to \cite{armagan2020measuring,Ge_2019_CVPR,spurr2020weakly}.
Below and in Table~\ref{table:supervision}, we compare our contribution with prior works.
\\
\textbf{Model-based Hand Pose and Shape Estimation.}\quad
Many hand models have been proposed to approximate hand shape via a parametric model \cite{ballan2012motion, khamis2015learning,romero2017embodied,tkach2016sphere}.
%
In this paper, we employ a hand model named MANO \cite{romero2017embodied} that maps pose and shape parameters to a triangle mesh \cite{boukhayma20193d,chen2020joint,hasson2019learning, zimmermann2019freihand}.

Because the parametric model contains abundant structure priors of human hands, recent works integrate hand model as a differentiable layers in neural networks \cite{Baek_2019_CVPR,Baek_2020_CVPR,boukhayma20193d,hasson2020leveraging,hasson2019learning, wang_SIGAsia2020,zhou2020monocular,zimmermann2019freihand}.
Among them, \cite{Baek_2019_CVPR,wang_SIGAsia2020, zhou2020monocular} output a set of intermediate estimations, like segmentation mask and 2D keypoints, and then maps these representations to the MANO parameters.
Different from them, we aim at demonstrating the feasibility of a self-supervised framework using an intuitive autoencoder. We additionally output 2D keypoint estimation from another branch and use it only during training to facilitate 3D reconstruction.
%
More generally, recent methods \cite{Baek_2020_CVPR, boukhayma20193d,hasson2020leveraging,hasson2019learning, zimmermann2019freihand} directly adopt an autoencoder that couples an image feature encoding stage with a model-based decoding stage.
Unlike \cite{hasson2020leveraging,hasson2019learning}, we focus on hand recovery and do not use any annotation about objects.
More importantly, the above methods use 3D annotations as supervision, while the proposed method does not rely on any ground truth annotations.
\\
\textbf{3D Hand Pose and Shape Estimation with Limited Supervision.}\quad
2D annotation is cheaper than 3D annotation, but it is difficult to deal with the ambiguity of depth and scale.
\cite{cai2018weakly} use a depth map to perform additional weak supervision to strengthen 2D supervision.
\cite{spurr2020weakly} proposes biomechanical constraints to help the network output feasible 3D hand configurations.
\cite{panteleris2018using} detects 2D hand keypoints and directly fits a hand model to the 2D detection.
\cite{Kulon_2020_CVPR} gathers a large-scale dataset through an automated data collection method similar to \cite{panteleris2018using} and then uses the collected mesh as supervision.
In this work, we limit biomechanical feasibility by introducing a set of constraints on the skin model instead of only impose constraints on the skeleton as~\cite{spurr2020weakly}.
In contrast to \cite{cai2018weakly, Kulon_2020_CVPR}, our method is designed to verify the feasibility of (noisy) 2D supervision and do not introduce any extra 2.5D or 3D data. 
\\
\textbf{Self-supervised 3D Reconstruction.}\quad Recently, there are methods that propose to learn 3D geometry from monocular image only. 
For example, \cite{Wu_2020_CVPR} proposes an unsupervised approach to learn 3D deformable objects from raw single-view images, but they assume the object is perfectly symmetric, which is not the case in the hand reconstruction. 
\cite{goel2020shape} removes out keypoints from supervision signals, but it uses ground truth 2D silhouette as supervision and only deals with categories with small intra-class shape differences, such as birds, shoes, and cars. 
\cite{wan2019self} proposes a depth-based self-supervised 3D hand pose estimation method, but the depth image provides much more strong evidence and supervision than the RGB image.
Recently, \cite{chen2020self, tewari2018self, tewari2017mofa} propose self-supervised face reconstruction with the use of 3D morphable model of face (3DMM) \cite{blanz1999morphable} and 2D landmarks detection.
Our approach is similar to them, but the hand is relatively non-flat and asymmetrical when compared with the 3D face, and the hand suffers from more severe self-occlusion. These characteristics make this self-supervised hand reconstruction task more challenging.\\
\textbf{Texture Modeling in Hand Recovery.}\quad 
\cite{de2011model, de2008model} exploit shading and texture information to handle the self-occlusion in the hand tracking system. Recently, \cite{qian2020parametric} uses principal component analysis (PCA) to build a parametric texture model of hand from a set of textured scans. In this work, we try to model texture from self-supervised training without introducing extra data, and further investigate whether the texture modeling helps with the shape modeling.

From the above analysis and comparison of related work, we believe that self-supervised 3D hand reconstruction is feasible and significant, but to the best of our knowledge, no such idea has been studied in this field.
In this work, we fill this gap and propose the first self-supervised 3D hand reconstruction network, and prove the effectiveness of the proposed method through experiments.
\section{Method}
Our method enables end-to-end learning of a 3D hand reconstruction network in a self-supervised manner, as illustrated in Fig.~\ref{fig:pipeline}.
To this end, we use an autoencoder that receives an image of a hand as input and outputs hand pose, shape, texture, and camera viewpoint (Section~\ref{sec:encoding} and \ref{sec:decoding}). 
We generate multiple 2D representations in image space (Section~\ref{sec:2D}) and design a series of loss functions and regularization terms for network training (Section~\ref{sec:losses}).
In the following, we describe the proposed method in detail.
\subsection{Deep Hand Encoding}\label{sec:encoding}
Given a image $I$ containing a hand, the network first uses an EfficientNet-b0 backbone \cite{tan2019efficientnet} to encode the image into a geometry semantic code vector $x$ and a texture semantic code vector $y$. The geometry semantic code vector $x$ parameterizes the hand pose $\theta \in \mathbb{R}^{30}$, shape $\beta \in \mathbb{R}^{10}$, scale $s \in \mathbb{R}^{1}$, rotation $R \in \mathbb{R}^{3}$ and translation $T \in \mathbb{R}^{3}$ in a unified manner: $x = (\theta, \beta, s, R, T)$. The texture semantic code vector $y$ parameterizes the hand texture $C \in \mathbb{R}^{778\times3}$ and scene lighting $L \in \mathbb{R}^{11}$ in a unified manner: $y = (C, L)$.
\subsection{Model-based Hand Decoding}\label{sec:decoding}
Given the geometry semantic code vector $x$ and the texture semantic code vector $y$, our model-based decoder generates a textured 3D hand model in the camera space.
In the following, we describe the used hand model and decoding network in detail.\\
\textbf{Pose and Shape Representation.}\quad The hand surface is represented by a manifold triangle mesh $M\equiv(V, F)$ with $n=778$ vertices $V=\lbrace v_{i}\in \mathbb{R}^{3} \vert 1\le i \le n\rbrace$ and faces $F$. 
The faces $F$ indicates the connection of the vertices in the hand surface, where we assume the face topology keeps fixed.
Given the mesh topology, a set of $k=21$ joints $J=\lbrace j_{i}\in \mathbb{R}^{3} \vert 1\le i \le k\rbrace$ can be directly formulated from the hand mesh. 
Here, the hand mesh and joints are recovered from the pose vector $\theta$ and the shape vector $\beta$ via MANO which is a low-dimensional parametric model learned from more than two thousand 3D hand scans \cite{romero2017embodied}.\\
\textbf{3D Hand in Camera Space.}\quad After representing 3D hand via MANO hand model from pose and shape parameters, the mesh and joints are located in the hand-relative coordinate systems. 
To represent the output joints and mesh in the camera coordinate system, we use the estimated scale, rotation and translation to conserve the original hand mesh $M_{0}$ and joints $J_{0}$ into the final representations: $M = s{M}_{0}R + T$ and $J = s{J}_{0}R + T$.
\\
\textbf{Texture and Lighting Representation.}\quad We use per-vertex RGB value of $n=778$ vertices to represent the texture of hand $C = \lbrace c_{i} \in \mathbb{R}^{3}| 1\le i \le n \rbrace$, where $c_{i}$ yields the RGB values of vertex $i$. 
In our model, we use a simple ambient light and a directional light to simulate lighting conditions \cite{kato2018neural}.
The lighting vector $L$ parameterizes ambient light intensity $l^a \in \mathbb{R}^{1}$, ambient light color $l^a_c \in \mathbb{R}^{3}$, directional light intensity color $l^d \in \mathbb{R}^{1}$, directional light color $l^d_c \in \mathbb{R}^{3}$, and directional light direction $n^d \in \mathbb{R}^{3}$ in a unified representation: $L = (l^a, l^a_c, l^d, l^d_c, n^d)$.

\subsection{Represent Hand in 2D}\label{sec:2D}
A set of estimated 3D joints within the camera scope can be projected into the image space by camera projection. 
Similarly, the output textured model can be formulated into a realistic 2D hand image through a neural renderer. 
In addition to the 2D keypoints projected from the model-based 3D joints, we can also estimate the 2D position of each keypoint in the input image. 
Here, we represent 2D hand in three modes and explore the complementarity among them.\\
\textbf{Joints Projection.}\quad Given a set of 3D joints in camera coordinates $J$ and the intrinsic parameters of the camera, we use camera projection $\Pi$ to project 3D joints into a set of $k=21$ 2D joints $J^{pro}=\lbrace j^{pro}_i \in \mathbb{R}^{2} \vert 1\le i \le k\rbrace$, where $j^{pro}_i$ yields the position of the $i$-th joint in image UV coordinates: $J^{pro} = \Pi (J)$. 
\\
\textbf{Image Formation.}\quad A 3D mesh renderer is used to conserve the triangle hand mesh into a 2D image, here we use an implementation\footnote{\scriptsize \url{https://github.com/daniilidis-group/neural\_renderer}} of \cite{kato2018neural}.
Given the 3D mesh $M$, the texture of the mesh $C$ and the lighting $L$, the neural renderer~$\Delta$ can generate a silhouette of hand $S^{re}$ and a color image $I^{re}$: $S^{re},I^{re}  = \Delta (M,C,L)$.\\
\textbf{Extra 2D Joint Estimation.}\quad 
Projecting model-based 3D joints into 2D helps the projected 2D keypoints retain structural information, but at the same time gives up the independence of each key point.
In view of this matter, we additionally use a 2D keypoint estimator to directly estimate a set of $k=21$ independent 2D joints \mbox{$J^{2d} =\lbrace j^{2d}_i \in \mathbb{R}^{2} \vert 1\le i \le k\rbrace$}, where $j^{2d}_i$ indicates the position of the $i$-th joint in image UV coordinates. In our 2D keypoint estimator, a stacked hourglass network \cite{newell2016stacked} along with an integral pose regression \cite{sun2018integral} is used. Note that the 2D hand pose estimation module is optionally deployed in the training period and is not required during the inference.
\subsection{Training Objective}\label{sec:losses}
Our overall training loss $E$ consists of three parts including 3D branch loss $E_{3d}$, 2D branch loss $E_{2d}$, and 2D-3D consistency loss $E_{con}$: 
\begin{equation}E= w_{3d}E_{3d} + w_{2d}E_{2d} + w_{con}E_{con}\end{equation}
Note, $E_{2d}$ and $E_{con}$ are optional and only used when the 2D estimator is applied. The constant weights $w_{3d}$, $w_{2d}$ and $w_{con}$ balance the three terms.
In the following, we describe these loss terms in detail. 
\subsubsection{Losses of the 3D Branch}
To train the model-based 3D hand decoder, we enforce geometric alignment $E_{geo}$, photometric alignment $E_{photo}$, and statistical regularization $E_{regu}$:
\begin{equation}
    E_{3d}  = w_{geo}E_{geo} + w_{photo}E_{photo} + w_{regu}E_{regu}
\end{equation}
\textbf{Geometric Alignment.} We propose a geometric alignment loss $E_{geo}$ based on the detected 2D keypoints which are obtained at an offline stage through an implementation\footnote{\scriptsize \url{https://github.com/Hzzone/pytorch-openpose}} of~\cite{cao2019openpose}.
The detected 2D keypoints $L =\lbrace (j^{de}_{i}, {con}_{i}) \vert 1\le i \le k \rbrace$ allocate each keypoint with a 2D position $j^{de}_{i} \in \mathbb{R}^{2}$ and a 1D confidence ${con}_{i} \in [0,1]$.
The geometric alignment loss in the 2D image space consists of a joint location loss $E_{loc}$ and a bone orientation loss $E_{ori}$.
The joint location loss $E_{loc}$ enforces the projected 2D keypoints $J^{pro}$ to be close to its corresponding 2D detections $J^{de}$, and the bone orientation loss $E_{ori}$ enforces the $m=20$ bones of these two sets of keypoints to be aligned.
\begin{equation}
    E_{loc}=\frac{1}{k}\sum_{i=1}^{k}{{con}_i}\mathcal{L}_{SmoothL1}(j^{de}_{i},j^{pro}_{i})
\label{eq:lmloc}\end{equation}
\begin{equation}
    E_{ori}=\frac{1}{m}\sum_{i=1}^{m}{{con}^{bone}_i}{\parallel \nu^{de}_{i}-\nu^{pro}_{i}\parallel}^2_2
\label{eq:lmori}
\end{equation}
Here, a SmoothL1 loss \cite{huber1992robust} is used in Eq.~\ref{eq:lmloc} to make the loss term to be more robust to local adjustment since the detection keypoints are not fit well with the MANO keypoints.
In~Eq.~\ref{eq:lmori}, $\nu^{de}_{i}$ and $\nu^{pro}_{i}$ are the normalized $i$-th bone vector of the detected 2D joints and the projected 2D joints, respectively, and ${con}^{bone}_i$ is the product of the confidence of the two detected 2D joints of the $i$-th bone. The overall geometric alignment loss $E_{geo}$ is the weighted sum of $E_{loc}$ and $E_{ori}$ with a weighting factor $w_{ori}$:
\begin{equation}
    E_{geo} = E_{loc} + w_{ori}E_{ori}
\end{equation}
\textbf{Photometric Consistency.}\quad For the image formation, the ideal result is the rendered color image $I^{re}$ matches the foreground hand of the input $I$. To this end, we employ a photometric consistency which has two parts, the \textit{pixel loss} ${E}_{pixel}$ is computed by averaging the least absolute deviation (L1) distance for all visible pixels to measure the pixel-wise difference, and the structural similarity (SSIM) loss ${E}_{SSIM}$ is the structural similarity between the two images \cite{wang2004image}.
\begin{equation}
\label{eq:pho}
{E}_{pixel}=\frac{{con}_{sum}}{\mid S^{re}\mid}\sum_{(u,v)\in S^{re}}{\parallel I_{u,v}-I^{re}_{u,v} \parallel}_2
\end{equation}
\begin{equation}
{E}_{SSIM}=1-SSIM({I}\odot{S^{re}},I^{re})
\end{equation}
Here, the rendered silhouette $S^{re}$ is used to get the foreground part of the input image for loss computation. 
In Eq.~\ref{eq:pho}, we use ${con}_{sum}$, which is the sum of the detection confidence of all keypoints, to distinguish different training samples.
This is because we think that low-confidence samples correspond to ambiguous texture confidence, e.g., the detection confidence of an occluded hand is usually low. 
The photometric consistency loss ${E}_{photo}$ is the weighted sum of ${E}_{pixel}$ and ${E}_{SSIM}$ by a weighting factor $w_{SSIM}$.
\begin{equation}
{E}_{photo}={E}_{pixel}+w_{SSIM}{E}_{SSIM}
\end{equation}
\begin{figure}[tb]
    \setlength{\abovecaptionskip}{0.cm}
	\begin{center}
		\includegraphics[width=1\linewidth]{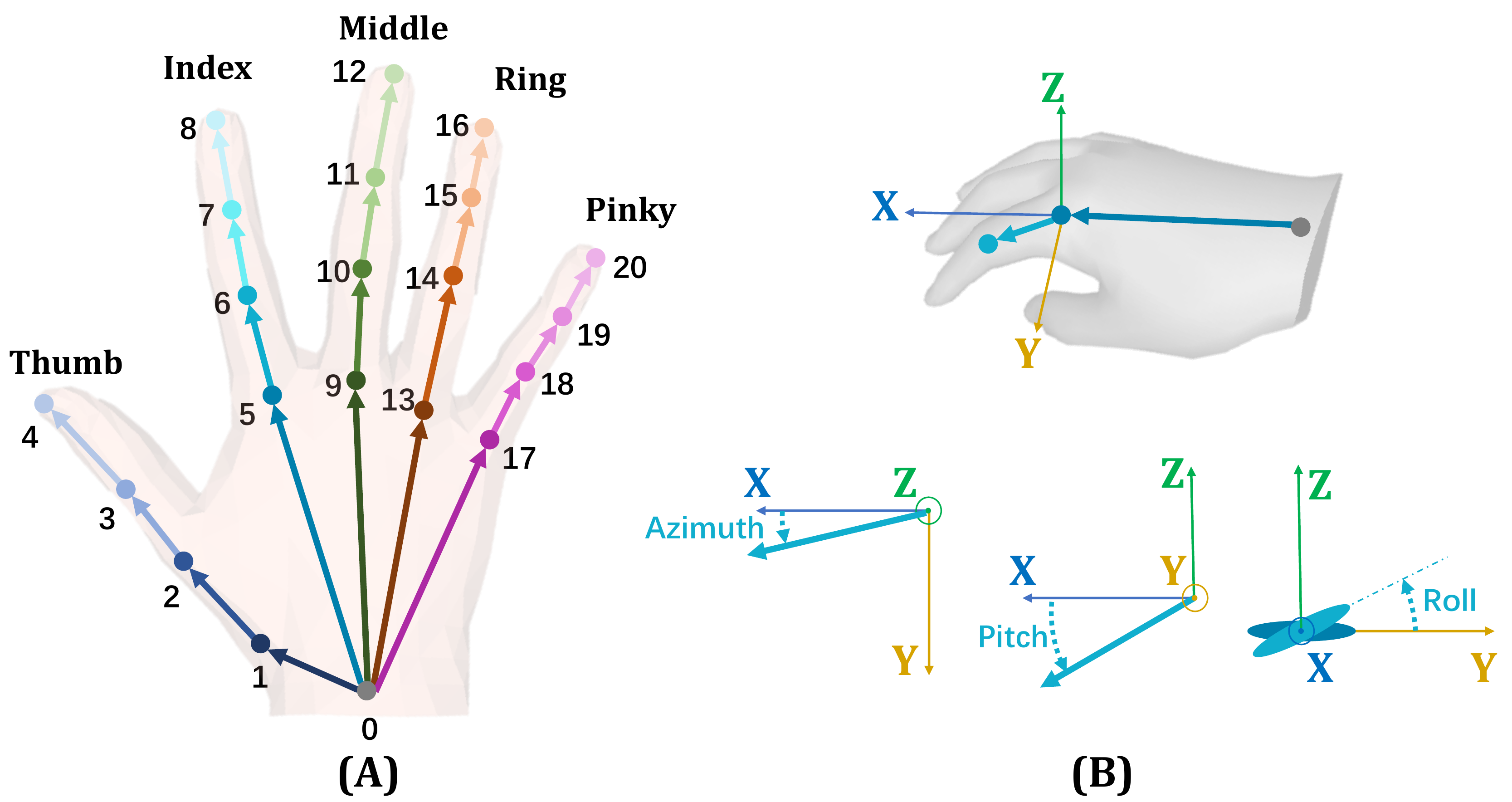}
	\end{center}
	\caption{(A) Joint skeleton structure. (B) A sample of bone rotation angles. The five bones ($\protect\overrightarrow{0\underline{1}}$,$\protect\overrightarrow{0\underline{5}}$,$\protect\overrightarrow{0\underline{9}}$,$\protect\overrightarrow{0\underline{13}}$,$\protect\overrightarrow{0\underline{17}}$) on the palm are fixed.
	Each finger has 3 bones, and the relative orientation of each bone from its root bone is represented by azimuth, pitch, and roll.}
\label{fig:prior}
\vspace{-0.1cm}
\end{figure}
\textbf{Statistical Regularization.}\quad 
During training, to make the results plausible, we introduce regularization terms, including shape regularization $E_{\beta}$, texture regularization $E_{C}$, scale regularization $E_{s}$, and 3D joints regularization $E_{J}$. The shape regularization term is defined as $E_{\beta}=\parallel \beta-\bar{\beta} \parallel$ to encourage the estimated hand model shape $\beta$ to be close to the average shape $\bar{\beta}= \vec{0} \in \mathbb{R}^{10}$.
The texture regularization $E_{C}$ is used to penalize outlier RGB values.
The scale regularization term $E_{s}$ is used to ensure the output hand is of appropriate size, so as to help determine the depth of the output in this monocular 3D reconstruction task.
As for the regularization constraints on skeleton $E_{J}$, we define feasible range for each rotation angle $a_i$ (as shown in Fig.~\ref{fig:prior}B) and penalize those who exceed the feasible threshold.
We provide more details about $E_{C}$,$E_{s}$ and $E_{J}$ in the Appendix.

The statistical regularization $E_{regu}$ is the weighted sum of $E_{\beta}$, $E_{C}$, ${E}_{s}$ and $E_{J}$ with weighting factors $w_{C}$, $w_{s}$ and $w_{J}$:
\begin{equation}
{E}_{regu}={E}_{\beta}+w_{C}{E}_{C}+w_{s}{E}_{s}+w_{J}{E}_{J}
\end{equation}
\subsubsection{2D-3D Consistency}
\textbf{Losses of the 2D Branch.}\quad For the 2D keypoint estimator, we use a joint location loss as Eq.~\ref{eq:lmloc} with replacing the projected 2D joint $j_i^{pro}$ by estimated 2D joint $j_i^{2d}$:
\begin{equation}
    E_{2d}=\frac{1}{k}\sum_{i=1}^{k}{{con}_i}\mathcal{L}_{SmoothL1}(j^{de}_{i},j^{2d}_{i})
\label{eq:lm2d}\end{equation}
\textbf{2D-3D Consistency Loss.}\quad Since outputs of the 2D branch and the 3D branch are intended to represent the same hand in different spaces, they should be consistent when they are transferred to the same domain.
Through this consistency, structural information contained in the 3D reconstruction network can be introduced into the 2D keypoint estimator, and meanwhile estimated 2D keypoints can provide joint-wise geometric cues for 3D hand reconstruction.
To this end, we propose a novel 2D-3D consistency loss to link per projected 2D joint $j_i^{pro}$ with its corresponding estimated 2D joint $j_{i}^{2d}$:
\begin{equation}
    E_{con}=\frac{1}{k}\sum_{i=1}^{k}\mathcal{L}_{SmoothL1}(j^{pro}_{i},j^{2d}_{i})
\label{eq:cons}\end{equation}
\begin{table*}[tb]
\vspace{-0.5cm}
\centering
\makebox[0pt][c]{\parbox{1.05\textwidth}{
\begin{minipage}[h]{0.64\textwidth}
    \centering
    \begin{minipage}[t]{0.9\textwidth}
        \centering
        \setlength{\abovecaptionskip}{0.cm}
        \scalebox{0.78}{
        \begin{tabular}{cccccccc}
        \Xhline{1pt}
        Supervision & Method & $\rm {AUC}_{J}$$\uparrow$ & MPJPE$\downarrow$&$\rm {AUC}_{V}$$\uparrow$ &  MPVPE$\downarrow$ & ${\rm F}_5$$\uparrow$ & ${\rm F}_{15}$$\uparrow$\\
        \hline
        \multirow{4}{*}{3D} &\cite{zimmermann2019freihand}(2019)& 0.35 & 3.50 & 0.74 & 1.32 & 0.43 & 0.90  \\
        &\cite{hasson2019learning}(2019) & 0.74 & 1.33 & 0.74 & 1.33 & 0.43 & 0.91\\
        &\cite{boukhayma20193d}(2019) & \textbf{0.78} & \textbf{1.10} & \textbf{0.78} & \textbf{1.09} & \textbf{0.52} & \textbf{0.93}\\
        &\cite{qian2020parametric}(2020) & \textbf{0.78} & \underline{1.11} & \textbf{0.78}& \underline{1.10} & \underline{0.51} & \textbf{0.93}\\
        \hline
        \multirow{1}{*}{2D} &\cite{spurr2020weakly}(2020)* & \textbf{0.78} & 1.13 & - & - & - & -\\
        \hline
        - & Ours & \underline{0.77} & 1.18 & \underline{0.77} & 1.19 & 0.48 & \underline{0.92}\\
        \Xhline{1pt}
        \end{tabular}}
        \vspace{0.1cm}
        \caption{Comparison of main results on the FreiHAND test set. The performance of our self-supervised method is comparable to the recent fully-supervised and weakly-supervised methods. \cite{spurr2020weakly}* also uses synthetic training data with 3D supervision. 
        }
        \label{table:sc-ssl}
    \end{minipage}%
    \vspace{0.2cm}
    \par
    \begin{minipage}[t]{0.9\textwidth}
        \centering
        \setlength{\abovecaptionskip}{0.cm}
        \scalebox{0.78}{
        \begin{tabular}{cccccccc}
        \Xhline{1pt}
        Supervision & Method & $\rm {AUC}_{J}$$\uparrow$ & MPJPE$\downarrow$& $\rm {AUC}_{V}$$\uparrow$ & MPVPE$\downarrow$ & ${\rm F}_5$$\uparrow$ & ${\rm F}_{15}$$\uparrow$\\
        \hline
        \multirow{3}{*}{3D} 
        &\cite{hasson2019learning}(2019) & - & - & - & 1.30 & 0.42 & 0.90\\
        &\cite{hampali2020honnotate}(2020) & - & - & - & \textbf{1.06} & \textbf{0.51} & \textbf{0.94} \\
        &\cite{hasson2020leveraging}(2020) & \textbf{0.773} & \textbf{1.11} & \underline{0.773} & 1.14 & 0.43 & \underline{0.93} \\
        \hline
        - & Ours & \textbf{0.773} & \underline{1.14} & \textbf{0.777} & \underline{1.12} & \underline{0.45} & \underline{0.93}\\
        \Xhline{1pt}
        \end{tabular}}
        \vspace{0.1cm}
        \caption{Comparison of main results on the HO-3D test set. Without using any object information and hand annotation, our hand pose and shape estimation method performs comparably with recent fully-supervised methods.
        }
        \label{table:sc-ssl-ho}
    \end{minipage}%
\end{minipage}
\begin{minipage}[h]{0.40\textwidth}
	\centering
	\includegraphics[width=1.06\linewidth]{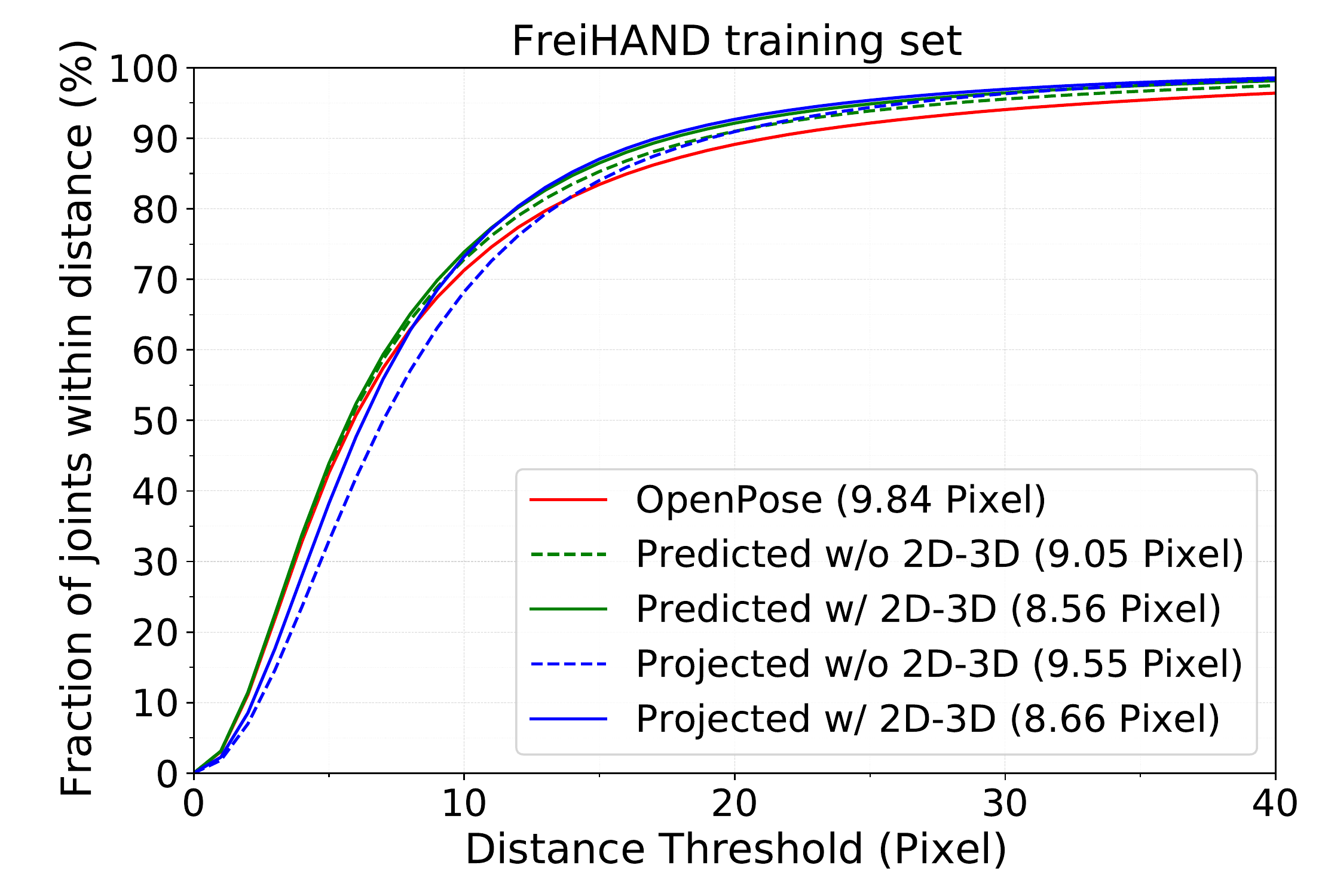}
	\captionof{figure}{A comparison of 2D keypoint sets used or outputted at the training stage on \mbox{FreiHAND}. The fraction of joints within distance is plotted. The average 2D distances in \textit{pixel} are shown in the legend. Refer to \mbox{Section~\ref{2d_comp}} for details.}
	\label{fig:2d}
\end{minipage}
}}
\vspace{-0.3cm}
\end{table*}
\section{Experiments}
In this section, we first present datasets and evaluation metrics (Section~\ref{sec:datasets}), and implementation details \mbox{(Section~\ref{sec:implementation})}. Then, we show the performance of our method and conduct comprehensive analysis (Section~\ref{sec:evaluation} and \ref{sec:ablation_study}).
\subsection{Datasets and evaluation metrics}\label{sec:datasets}
We evaluated our method on two challenging real datasets, both of which evaluate 3D joints and 3D meshes. The results are reported results through online submission systems\footnote{\scriptsize \url{https://competitions.codalab.org/competitions/21238}}${}^{,}$\footnote{\scriptsize \url{https://competitions.codalab.org/competitions/22485}}.\\
\textbf{FreiHAND.}\quad The FreiHAND dataset is a large-scale real-world dataset,
which contains 32,560 training samples and 3,960 test samples. For each training sample, one real RGB image and extra three images with different synthetic backgrounds are provided.
Part of the sample is a hand grabbing an object, but it does not provide any annotations for the foreground object, which poses additional challenges.\\
\textbf{HO-3D.}\quad The HO-3D dataset collects color images of a hand interacting with an object. 
The dataset is made of 68 sequences, totaling 77,558 frames of 10 users manipulating one among 10 different objects. 
The training set contains 66,034 images and the test set contains 11,524 images. 
The objects in this dataset are larger than that in FreiHAND, thus resulting in larger occlusions to hands. 
\newline \textbf{Evaluation Metrics.}\quad 
We evaluate 3D hand reconstruction by evaluating 3D joints and 3D meshes. For 3D joints, we report the \textbf{mean per joint position error} (MPJPE) in the Euclidean space for all joints on all test frames in \textit{cm} and the \textbf{area under the curve} (AUC) of the PCK $\rm AUC_{J}$. 
Here, the PCK refers to the percentage of correct keypoints, is plotted using 100 equally spaced thresholds between 0\textit{mm} to 50\textit{mm}.
For 3D meshes, we report the \textbf{mean per vertex position error} (MPVPE) in the Euclidean space for all joints on all test frames in \textit{cm} and the AUC of the percentage of correct vertex $\rm AUC_{V}$. 
We also compare the F-score \cite{knapitsch2017tanks} which is the harmonic mean of recall and precision for a given distance threshold. We report distance threshold at 5\textit{mm} and 15\textit{mm} and report F-score of mesh vertices at 5\textit{mm} and 15\textit{mm} by ${\rm F}_5$ and ${\rm F}_{15}$. 
Following the previous works \cite{hampali2020honnotate,zimmermann2019freihand}, we compare aligned prediction results with Procrustes alignment, and all 3D results are evaluated by the online evaluation system. 
For 2D joints, we report the MPJPE in \textit{pixel} and the curve plot of \textbf{fraction of joints within distance}.

\subsection{Implementation}\label{sec:implementation}
Pytorch \cite{paszke2017automatic} is used for implementation. For the 3D reconstruction network, the EfficientNet-b0 \cite{tan2019efficientnet} is pre-trained on the ImageNet dataset. The 2D keypoint estimator along with the 2D-3D consistency loss is optionally used. If we train the whole network with the 2D keypoint estimator, a stage-wise training scheme is used.
We train the 2D keypoint estimator and 3D reconstruction network by 90 epochs separately, where $E_{3d}$ and $E_{2d}$ are used, respectively. The initial learning rate is ${10}^{-3}$ and reduced by a factor of 2 after every 30 epochs.
Then we finetune the whole network with $E$ by 60 epochs with the learning rate initialized to $2.5\times {10}^{-4}$ and reduced by a factor of 3 after every 20 epochs. We use Adam \cite{kingma2014adam} to optimize the network weights with a batch size of 64. We train our model on two NVIDIA Tesla V100 GPUs, which takes around 36 hours for training on FreiHAND. We provide more details in the Appendix.

\subsection{Comparison with State-of-the-art Methods}\label{sec:evaluation}
We give comparison on FreiHAND with four recent model-based fully-supervised methods \cite{boukhayma20193d,hasson2019learning, qian2020parametric,zimmermann2019freihand} and a state-of-the-art weakly-supervised method \cite{spurr2020weakly} in \mbox{Table~\ref{table:sc-ssl}}.
Note that \cite{moon2020i2l} is not included here since it designs an advanced ``image-to-lixel" prediction instead of directly regress MANO parameters.
Our approach focuses on providing a self-supervised framework with lightweight components, where the hand regression scheme is still affected by highly non-linear mapping.
Therefore, we make a fairer comparison with popular model-based methods \cite{boukhayma20193d,hasson2019learning,qian2020parametric,zimmermann2019freihand} to demonstrate the performance of this self-supervised approach.
Without using any annotation, our approach outperforms \cite{hasson2019learning,zimmermann2019freihand} on all evaluation metrics and achieves comparable performance to \cite{boukhayma20193d,qian2020parametric}.
\cite{spurr2020weakly} only outputs 3D pose, and its pose performance is slightly better than our results on FreiHAND test set but with much more training data used including RHD dataset \cite{zimmermann2017learning} (with 40,000+ synthetic images and 3D annotations) as well as 2D ground truth annotation of the FreiHAND.

In the hand-object interaction scenario, we compare with three recent fully-supervised methods on HO-3D in Table~\ref{table:sc-ssl-ho}.
Compared to the hand branch of \cite{hasson2019learning}, our self-supervised results show higher mesh reconstruction performance where we get a 14\% reduction in MPVPE.
Compared with \cite{hasson2020leveraging}, which is a fully-supervised joint hand-object pose estimation method, our approach obtains comparable joints and shape estimation results.
\cite{hampali2020honnotate} gets slightly better shape estimation results than ours, which may be due to it uses multi-frame joint hand-object pose refinement and mesh supervision.

In Fig.~\ref{fig:qualitative}, we show 2D keypoint detection from \mbox{OpenPose~\cite{cao2019openpose}} and our hand reconstruction results of difficult samples. We also compare the reconstruction results with MANO-CNN, which directly estimates MANO parameters with a CNN \cite{zimmermann2019freihand}, but we modify its backbone to be the same as ours. Our results are more accurate and additionally with texture.
\begin{table*}[tb]
\vspace{-0.25cm}
\centering
\makebox[0pt][c]{\parbox{1\textwidth}{
\begin{minipage}[h]{0.62\textwidth}
    \setlength{\abovecaptionskip}{0.cm}
    \centering
    \scalebox{0.7}{
    \begin{tabular}{cccccccccc}
    \Xhline{1pt}
    \multicolumn{4}{c}{Losses} &
    \multirow{2}{*}{MPJPE$\downarrow$}&
    \multirow{2}{*}{MPVPE$\downarrow$} &
    \multirow{2}{*}{$\rm {AUC}_{J}$$\uparrow$} &  \multirow{2}{*}{$\rm {AUC}_{V}$$\uparrow$} & \multirow{2}{*}{${\rm F}_5$$\uparrow$} & \multirow{2}{*}{${\rm F}_{15}$$\uparrow$} \\
    \cline{1-4}
    ${E}_{loc}$, ${E}_{regu}$ & ${E}_{ori}$ & ${E}_{2d}$, ${E}_{con}$ & ${E}_{photo}$\\
    \hline
    $\checkmark$ & & & & 1.54 & 1.58 & 0.696 & 0.687 & 0.387 & 0.852 \\
    $\checkmark$ & $\checkmark$ & & & 1.24 & 1.26 & 0.754 & 0.750 & 0.457 & 0.903\\
    $\checkmark$ & $\checkmark$ & $\checkmark$ & ~ & 1.19& 1.20 & 0.764 & 0.763 & 0.479 & 0.915\\
    $\checkmark$ & $\checkmark$ & $\checkmark$ & $\checkmark$ & \textbf{1.18} & \textbf{1.19} & \textbf{0.766} & \textbf{0.765} & \textbf{0.483} & \textbf{0.917}\\
    \Xhline{1pt}
    \end{tabular}
    }
    \vspace{0.1cm}
    \caption{Ablation studies for different losses used in our method on the {FreiHAND} testing set.  Refer to Section~\ref{component} for details. 
    }
    \label{table:ablation}
\end{minipage}
\hfill
\begin{minipage}[h]{0.35\textwidth}
    \centering
    \setlength{\abovecaptionskip}{0.cm}
    \centering
    \scalebox{0.7}{
    \begin{tabular}{cccccccc}
    \Xhline{1pt}
    Dataset & Method & $\rm AUC_{J}$$\uparrow$ & $\rm AUC_{V}$$\uparrow$ & $\rm F_5$$\uparrow$ & $\rm F_{15}$$\uparrow$\\
    \hline
    \multirow{2}{*}{FreiHAND} & WSL & 0.730 & 0.725 & 0.42 & 0.89 \\
    &SSL & \textbf{0.766} & \textbf{0.765} & \textbf{0.48} & \textbf{0.92} \\
    \hline
    \multirow{2}{*}{HO-3D} & WSL & 0.765 & 0.769  & 0.44 & \textbf{0.93}  \\
    &SSL & \textbf{0.773} & \textbf{0.777}  & \textbf{0.45} & \textbf{0.93} \\
    \Xhline{1pt}
    \end{tabular}}
    \vspace{0.1cm}
    \caption{Comparison of self-supervised results and weakly-supervised results. Refer to Section~\ref{wsl-ssl} for details.}
    \label{table:ssl-wsl}
\end{minipage}
}}
\vspace{-0.1in}
\end{table*}
\begin{figure}[tb]
    \setlength{\abovecaptionskip}{0.cm}
	\includegraphics[width=1\linewidth]{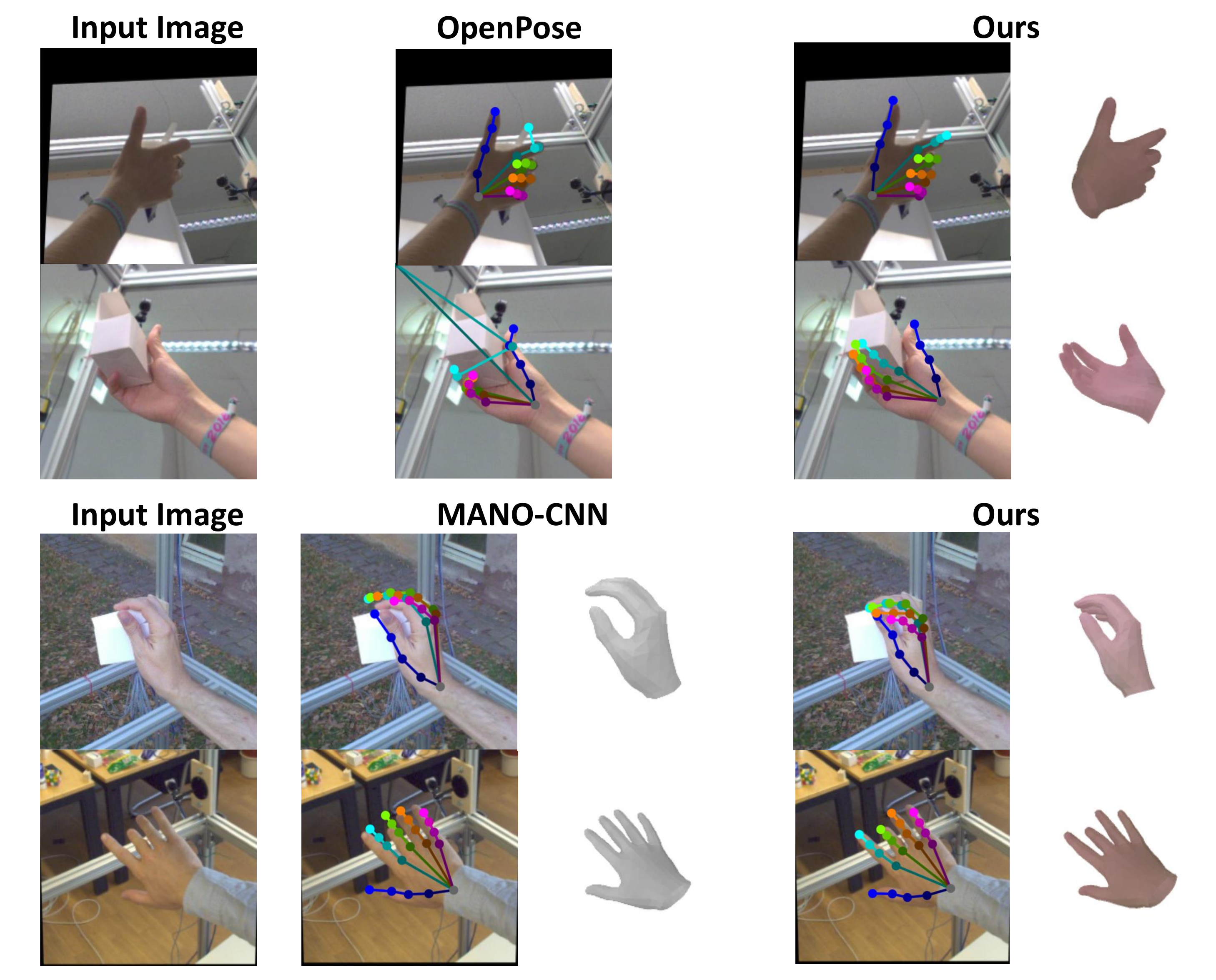}
	\caption{Qualitative comparison to OpenPose \cite{cao2019openpose} and MANO-CNN on the FreiHAND test set. For OpenPose, we visualize detected 2D keypoints. For our method and MANO-CNN, we visualize projected 2D keypoints and 3D mesh.
	}
	\label{fig:qualitative}
\vspace{-0.3cm}
\end{figure}
\subsection{Ablation Study}\label{sec:ablation_study}
\subsubsection{Effect of Each Component}
\label{component}
As presented in Table~\ref{table:ablation}, we give evaluation results on \mbox{FreiHAND} of settings with different components along with corresponding loss terms used in the network.
The baseline only uses the 3D branch with $E_{loc}$ and $E_{regu}$, then we add $E_{ori}$ which helps the MPJPE and MPVPE decrease by 19.5\%.
After adding the 2D branch with $E_{2d}$ and the 2D-3D consistency constrain $E_{con}$, the MPJPE and MPVPE further reduce by 4\%.
The $E_{photo}$ slightly improves the pose and shape estimation results.
\vspace{-0.3cm}
\subsubsection{Comparison of Different 2D Keypoint Sets}
\label{2d_comp}
In our approach, there are three sets of 2D keypoints, including detected keypoints $J^{de}$, estimated 2D keypoints $J^{2d}$, and output projected keypoints $J^{pro}$, where $J^{de}$ is used as supervision terms while $J^{2d}$ and $J^{pro}$ are output items.
In our setting, we use multiple 2D representations to boost the final 3D hand reconstruction, so we do not advocate the novelty of 2D hand estimation, but compare 2D accuracy in the training set to demonstrate the effect of learning from noisy supervision and the benefits of the proposed 2D-3D consistency.

Although we use OpenPose outputs as the keypoint supervision source (see \textbf{\textit{OpenPose}} in Fig.~\ref{fig:2d}), we get lower overall 2D MPJPE when we pre-train the 2D and 3D branches separately (see \textbf{\textit{Predicted w/o 2D-3D}} and \textbf{\textit{Projected w/o 2D-3D}} in Fig.~\ref{fig:2d}). 
After finetuning these two branches with 2D-3D consistency, we find both of them gain additional benefits.
After the finetuning, the 2D branch (\textbf{\textit{Predicted w/ 2D-3D}}) gains 5.4\% reduction in 2D MPJPE and the 3D branch (\textbf{\textit{Projected w/ 2D-3D}}) gains 9.3\% reduction in 2D MPJPE. 
From the curves, we can see that 2D keypoint estimation (including OpenPose and our 2D branch) gets higher accuracy in small distance thresholds while the regression-based methods (\textbf{\textit{Projected w/o 2D-3D}}) get higher accuracy with larger distance threshold. 
From the curves, the proposed 2D-3D consistency can improve the 3D branch in all distance thresholds, which verifies the rationality of our network design.
\subsubsection{Comparison with GT 2D Supervision}
\label{wsl-ssl}
We compare the weak-supervised (WSL) scheme using ground truth annotations with our self-supervised (SSL) approach to investigate the ability of our method to handle noisy supervision sources.
Both settings use the same network structure and implementation, and WSL uses the ground truth 2D keypoint annotations whose keypoint confidences are set to be the same.
As shown in Table~\ref{table:ssl-wsl}, our SSL approach has better performance than WSL settings on both datasets.
We think this is because the detection confidence information is embedded into the proposed loss functions, which helps the network discriminate different accuracy in the noisy samples.
In addition, we find that the SSL method outperforms the WSL method in a smaller amplitude on HO-3D (by 1.0\%) than that on FreiHand (by 4.9\%). We think this is because the HO-3D contains more occluded hands, resulting in poor 2D detection results.
Therefore, we conclude that noisy 2D keypoints can supervise shape learning for the hand reconstruction task, while the quality of the unlabeled image also has a certain impact.

\section{Discussion}
\label{discussion}
\begin{figure}[tb]
    \centering
    \setlength{\abovecaptionskip}{0.cm}
    \centering
	\includegraphics[width=1.1\linewidth]{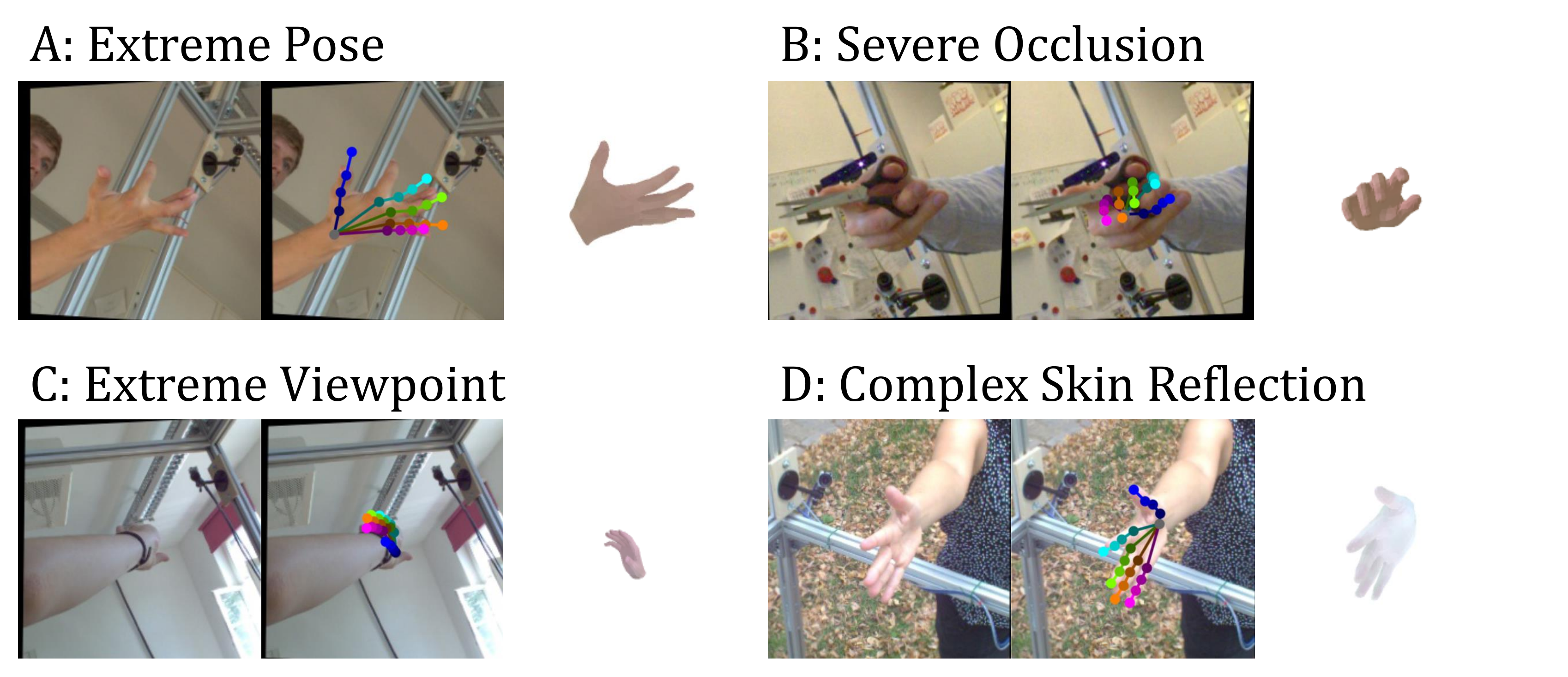}
	\caption{Failure cases. We show input image, projected joints, and 3D reconstruction. See Section~\ref{discussion} for details.
	}
	\label{fig:failure}
\vspace{-0.4cm}
\end{figure}
While our method results in accurate and vivid hand reconstruction in many challenging scenarios (e.g., hand-object interaction, self-occlusion), we also observe failure cases as shown in Fig.~\ref{fig:failure}. 
The reconstruction accuracy is lower in the extreme pose, severe occlusion, and extreme viewpoint, partly due to poor supervision from single-view 2D keypoint detection.
The texture modeling under complex skin reflections is inaccurate, which may be due to the fact that it is difficult for us to accurately simulate complex skin reflection using a simple illumination representation and a coarse hand mesh.
As shown in the last line of \mbox{Table~\ref{table:ablation}}, the texture modeling cannot bring a marked improvement to the shape reconstruction, which is also the case in \cite{qian2020parametric}.
This may be because the hand model \cite{romero2017embodied} is not meticulous enough, and the skin reflection simulation is not accurate.
\section{Conclusion}
We have presented a self-supervised 3D hand reconstruction network that can be trained from a collection of unlabeled hand images.
The network encodes the input image into a set of meaningful semantic parameters that represent hand pose, shape, texture, illumination, and the camera viewpoint, respectively.
These parameters can be decoded into a textured 3D hand mesh as well a set of 3D joints, and in turn, 3D mesh and joints can be projected into 2D image space, which enables our network to be end-to-end learned.
Our network performs well under noisy supervision sources from 2D hand keypoint detection while is able to obtain accurate 3D hand reconstruction from a single-view hand image.
Experimental results show that our method achieves comparable performance with state-of-the-art fully-supervised methods.
As for the future study, it is possible to extend the parametric hand mesh to other representations (e.g., signed distance function) for more detailed hand surface representation.
We also believe that more accurate skin reflection modeling can help hand reconstruction with higher fidelity.
\\
\\
{\small
\textbf{Acknowledgement}: 
This work is supported by the Fundamental Research Funds for the Central Universities No.2042020KF0016 and also supported in part by National Science Foundation Grant CNS-1951952.}

{\small
\normalem
\bibliographystyle{ieee_fullname}
\bibliography{CVPR21-arXiv}

\begin{thebibliography}{10}\itemsep=-1pt

\bibitem{armagan2020measuring}
Anil Armagan, Guillermo Garcia-Hernando, Seungryul Baek, Shreyas Hampali, Mahdi
  Rad, Zhaohui Zhang, Shipeng Xie, MingXiu Chen, Boshen Zhang, Fu Xiong, et~al.
\newblock Measuring generalisation to unseen viewpoints, articulations, shapes
  and objects for 3d hand pose estimation under hand-object interaction.
\newblock In {\em European Conference on Computer Vision}, 2020.

\bibitem{athitsos2003estimating}
Vassilis Athitsos and Stan Sclaroff.
\newblock Estimating 3d hand pose from a cluttered image.
\newblock In {\em Conference on Computer Vision and Pattern Recognition}, 2003.

\bibitem{Baek_2019_CVPR}
Seungryul Baek, Kwang~In Kim, and Tae-Kyun Kim.
\newblock Pushing the envelope for rgb-based dense 3d hand pose estimation via
  neural rendering.
\newblock In {\em Conference on Computer Vision and Pattern Recognition}, 2019.

\bibitem{Baek_2020_CVPR}
Seungryul Baek, Kwang~In Kim, and Tae-Kyun Kim.
\newblock Weakly-supervised domain adaptation via gan and mesh model for
  estimating 3d hand poses interacting objects.
\newblock In {\em Conference on Computer Vision and Pattern Recognition}, 2020.

\bibitem{ballan2012motion}
Luca Ballan, Aparna Taneja, J{\"u}rgen Gall, Luc Van~Gool, and Marc Pollefeys.
\newblock Motion capture of hands in action using discriminative salient
  points.
\newblock In {\em European Conference on Computer Vision}, 2012.

\bibitem{blanz1999morphable}
Volker Blanz and Thomas Vetter.
\newblock A morphable model for the synthesis of 3d faces.
\newblock In {\em Proceedings of the 26th annual conference on Computer
  graphics and interactive techniques}, pages 187--194, 1999.

\bibitem{boukhayma20193d}
Adnane Boukhayma, Rodrigo~de Bem, and Philip~HS Torr.
\newblock 3d hand shape and pose from images in the wild.
\newblock In {\em Conference on Computer Vision and Pattern Recognition}, pages
  10843--10852, 2019.

\bibitem{cai2018weakly}
Yujun Cai, Liuhao Ge, Jianfei Cai, and Junsong Yuan.
\newblock Weakly-supervised 3d hand pose estimation from monocular rgb images.
\newblock In {\em European Conference on Computer Vision}, 2018.

\bibitem{cao2019openpose}
Zhe Cao, Tomas Simon, Shih-En Wei, and Yaser Sheikh.
\newblock Openpose: Realtime multi-person 2d pose estimation using part
  affinity fields.
\newblock {\em IEEE Transactions on Pattern Analysis and Machine Intelligence},
  2019.

\bibitem{chen2019so}
Yujin Chen, Zhigang Tu, Liuhao Ge, Dejun Zhang, Ruizhi Chen, and Junsong Yuan.
\newblock So-handnet: Self-organizing network for 3d hand pose estimation with
  semi-supervised learning.
\newblock In {\em International Conference on Computer Vision}, 2019.

\bibitem{chen2020joint}
Yujin Chen, Zhigang Tu, Di Kang, Ruizhi Chen, Linchao Bao, Zhengyou Zhang, and
  Junsong Yuan.
\newblock Joint hand-object 3d reconstruction from a single image with
  cross-branch feature fusion.
\newblock {\em arXiv preprint arXiv:2006.15561}, 2020.

\bibitem{chen2020self}
Yajing Chen, Fanzi Wu, Zeyu Wang, Yibing Song, Yonggen Ling, and Linchao Bao.
\newblock Self-supervised learning of detailed 3d face reconstruction.
\newblock {\em IEEE Transactions on Image Processing}, 29:8696--8705, 2020.

\bibitem{de2011model}
Martin de La~Gorce, David~J Fleet, and Nikos Paragios.
\newblock Model-based 3d hand pose estimation from monocular video.
\newblock {\em IEEE Transactions on Pattern Analysis and Machine Intelligence},
  2011.

\bibitem{de2008model}
Martin de La~Gorce, Nikos Paragios, and David~J Fleet.
\newblock Model-based hand tracking with texture, shading and self-occlusions.
\newblock In {\em Conference on Computer Vision and Pattern Recognition}, 2008.

\bibitem{ge2016robust}
Liuhao Ge, Hui Liang, Junsong Yuan, and Daniel Thalmann.
\newblock Robust 3d hand pose estimation in single depth images: from
  single-view cnn to multi-view cnns.
\newblock In {\em Conference on Computer Vision and Pattern Recognition}, 2016.

\bibitem{Ge_2019_CVPR}
Liuhao Ge, Zhou Ren, Yuncheng Li, Zehao Xue, Yingying Wang, Jianfei Cai, and
  Junsong Yuan.
\newblock 3d hand shape and pose estimation from a single rgb image.
\newblock In {\em Conference on Computer Vision and Pattern Recognition}, 2019.

\bibitem{goel2020shape}
Shubham Goel, Angjoo Kanazawa, and Jitendra Malik.
\newblock Shape and viewpoint without keypoints.
\newblock In {\em European Conference on Computer Vision}, 2020.

\bibitem{hampali2020honnotate}
Shreyas Hampali, Mahdi Rad, Markus Oberweger, and Vincent Lepetit.
\newblock Honnotate: A method for 3d annotation of hand and object poses.
\newblock In {\em Conference on Computer Vision and Pattern Recognition}, 2020.

\bibitem{hasson2020leveraging}
Yana Hasson, Bugra Tekin, Federica Bogo, Ivan Laptev, Marc Pollefeys, and
  Cordelia Schmid.
\newblock Leveraging photometric consistency over time for sparsely supervised
  hand-object reconstruction.
\newblock In {\em Conference on Computer Vision and Pattern Recognition}, 2020.

\bibitem{hasson2019learning}
Yana Hasson, Gul Varol, Dimitrios Tzionas, Igor Kalevatykh, Michael~J Black,
  Ivan Laptev, and Cordelia Schmid.
\newblock Learning joint reconstruction of hands and manipulated objects.
\newblock In {\em Conference on Computer Vision and Pattern Recognition}, 2019.

\bibitem{holl2018efficient}
Markus H{\"o}ll, Markus Oberweger, Clemens Arth, and Vincent Lepetit.
\newblock Efficient physics-based implementation for realistic hand-object
  interaction in virtual reality.
\newblock In {\em IEEE Conference on Virtual Reality and 3D User Interfaces},
  2018.

\bibitem{huang2020hand}
Lin Huang, Jianchao Tan, Ji Liu, and Junsong Yuan.
\newblock Hand-transformer: Non-autoregressive structured modeling for 3d hand
  pose estimation.
\newblock In {\em European Conference on Computer Vision}, 2020.

\bibitem{huber1992robust}
Peter~J Huber.
\newblock Robust estimation of a location parameter.
\newblock In {\em Breakthroughs in statistics}, pages 492--518. Springer, 1992.

\bibitem{iqbal2018hand}
Umar Iqbal, Pavlo Molchanov, Thomas Breuel Juergen~Gall, and Jan Kautz.
\newblock Hand pose estimation via latent 2.5 d heatmap regression.
\newblock In {\em European Conference on Computer Vision}, 2018.

\bibitem{kato2018neural}
Hiroharu Kato, Yoshitaka Ushiku, and Tatsuya Harada.
\newblock Neural 3d mesh renderer.
\newblock In {\em Conference on Computer Vision and Pattern Recognition}, 2018.

\bibitem{khamis2015learning}
Sameh Khamis, Jonathan Taylor, Jamie Shotton, Cem Keskin, Shahram Izadi, and
  Andrew Fitzgibbon.
\newblock Learning an efficient model of hand shape variation from depth
  images.
\newblock In {\em Conference on Computer Vision and Pattern Recognition}, 2015.

\bibitem{kingma2014adam}
Diederik~P Kingma and Jimmy Ba.
\newblock Adam: A method for stochastic optimization.
\newblock In {\em International Conference for Learning Representations}, 2014.

\bibitem{knapitsch2017tanks}
Arno Knapitsch, Jaesik Park, Qian-Yi Zhou, and Vladlen Koltun.
\newblock Tanks and temples: Benchmarking large-scale scene reconstruction.
\newblock {\em ACM Transactions on Graphics}, 2017.

\bibitem{Kulon_2020_CVPR}
Dominik Kulon, Riza~Alp Guler, Iasonas Kokkinos, Michael~M. Bronstein, and
  Stefanos Zafeiriou.
\newblock Weakly-supervised mesh-convolutional hand reconstruction in the wild.
\newblock In {\em Conference on Computer Vision and Pattern Recognition}, 2020.

\bibitem{moon2020i2l}
Gyeongsik Moon and Kyoung~Mu Lee.
\newblock I2l-meshnet: Image-to-lixel prediction network for accurate 3d human
  pose and mesh estimation from a single rgb image.
\newblock In {\em European Conference on Computer Vision}, 2020.

\bibitem{newell2016stacked}
Alejandro Newell, Kaiyu Yang, and Jia Deng.
\newblock Stacked hourglass networks for human pose estimation.
\newblock In {\em European Conference on Computer Vision}, 2016.

\bibitem{panteleris2018using}
Paschalis Panteleris, Iason Oikonomidis, and Antonis Argyros.
\newblock Using a single rgb frame for real time 3d hand pose estimation in the
  wild.
\newblock In {\em Winter Conference on Applications of Computer Vision}, 2018.

\bibitem{parelli2020exploiting}
Maria Parelli, Katerina Papadimitriou, Gerasimos Potamianos, Georgios Pavlakos,
  and Petros Maragos.
\newblock Exploiting 3d hand pose estimation in deep learning-based sign
  language recognition from rgb videos.
\newblock In {\em European Conference on Computer Vision}. Springer, 2020.

\bibitem{paszke2017automatic}
Adam Paszke, Sam Gross, Soumith Chintala, Gregory Chanan, Edward Yang, Zachary
  DeVito, Zeming Lin, Alban Desmaison, Luca Antiga, and Adam Lerer.
\newblock Automatic differentiation in pytorch.
\newblock In {\em Neural Information Processing Systems Workshops}, 2017.

\bibitem{qian2020parametric}
Neng Qian, Jiayi Wang, Franziska Mueller, Florian Bernard, Vladislav Golyanik,
  and Christian Theobalt.
\newblock Parametric hand texture model for 3d hand reconstruction and
  personalization.
\newblock In {\em European Conference on Computer Vision}. Springer, 2020.

\bibitem{romero2017embodied}
Javier Romero, Dimitrios Tzionas, and Michael~J Black.
\newblock Embodied hands: Modeling and capturing hands and bodies together.
\newblock {\em ACM Transactions on Graphics}, 2017.

\bibitem{spurr2020weakly}
Adrian Spurr, Umar Iqbal, Pavlo Molchanov, Otmar Hilliges, and Jan Kautz.
\newblock Weakly supervised 3d hand pose estimation via biomechanical
  constraints.
\newblock In {\em European Conference on Computer Vision}, 2020.

\bibitem{sun2018integral}
Xiao Sun, Bin Xiao, Fangyin Wei, Shuang Liang, and Yichen Wei.
\newblock Integral human pose regression.
\newblock In {\em European Conference on Computer Vision}, 2018.

\bibitem{tan2019efficientnet}
Mingxing Tan and Quoc~V Le.
\newblock Efficientnet: Rethinking model scaling for convolutional neural
  networks.
\newblock In {\em International Conference on Machine Learning}, 2019.

\bibitem{tewari2018self}
Ayush Tewari, Michael Zollh{\"o}fer, Pablo Garrido, Florian Bernard, Hyeongwoo
  Kim, Patrick P{\'e}rez, and Christian Theobalt.
\newblock Self-supervised multi-level face model learning for monocular
  reconstruction at over 250 hz.
\newblock In {\em Conference on Computer Vision and Pattern Recognition}, 2018.

\bibitem{tewari2017mofa}
Ayush Tewari, Michael Zollhofer, Hyeongwoo Kim, Pablo Garrido, Florian Bernard,
  Patrick Perez, and Christian Theobalt.
\newblock Mofa: Model-based deep convolutional face autoencoder for
  unsupervised monocular reconstruction.
\newblock In {\em International Conference on Computer Vision Workshops}, 2017.

\bibitem{tkach2016sphere}
Anastasia Tkach, Mark Pauly, and Andrea Tagliasacchi.
\newblock Sphere-meshes for real-time hand modeling and tracking.
\newblock {\em ACM Transactions on Graphics}, 2016.

\bibitem{tu2019action}
Zhigang Tu, Hongyan Li, Dejun Zhang, Justin Dauwels, Baoxin Li, and Junsong
  Yuan.
\newblock Action-stage emphasized spatiotemporal vlad for video action
  recognition.
\newblock {\em IEEE Transactions on Image Processing}, 28(6):2799--2812, 2019.

\bibitem{wan2019self}
Chengde Wan, Thomas Probst, Luc~Van Gool, and Angela Yao.
\newblock Self-supervised 3d hand pose estimation through training by fitting.
\newblock In {\em Conference on Computer Vision and Pattern Recognition}, 2019.

\bibitem{wang_SIGAsia2020}
Jiayi Wang, Franziska Mueller, Florian Bernard, Suzanne Sorli, Oleksandr
  Sotnychenko, Neng Qian, Miguel~A. Otaduy, Dan Casas, and Christian Theobalt.
\newblock {RGB2Hands: Real-Time Tracking of 3D Hand Interactions from Monocular
  RGB Video}.
\newblock {\em ACM Transactions on Graphics (Proceedings of SIGGRAPH Asia)},
  2020.

\bibitem{wang2004image}
Zhou Wang, Alan~C Bovik, Hamid~R Sheikh, and Eero~P Simoncelli.
\newblock Image quality assessment: from error visibility to structural
  similarity.
\newblock {\em IEEE Transactions on Image Processing}, 2004.

\bibitem{Wu_2020_CVPR}
Shangzhe Wu, Christian Rupprecht, and Andrea Vedaldi.
\newblock Unsupervised learning of probably symmetric deformable 3d objects
  from images in the wild.
\newblock In {\em Conference on Computer Vision and Pattern Recognition}, 2020.

\bibitem{yang2020seqhand}
John Yang, Hyung~Jin Chang, Seungeui Lee, and Nojun Kwak.
\newblock Seqhand: Rgb-sequence-based 3d hand pose and shape estimation.
\newblock In {\em European Conference on Computer Vision}, 2020.

\bibitem{yuan2018depth}
Shanxin Yuan, Guillermo Garcia-Hernando, Bj{\"o}rn Stenger, Gyeongsik Moon, Ju
  Yong~Chang, Kyoung Mu~Lee, Pavlo Molchanov, Jan Kautz, Sina Honari, Liuhao
  Ge, et~al.
\newblock Depth-based 3d hand pose estimation: From current achievements to
  future goals.
\newblock In {\em Conference on Computer Vision and Pattern Recognition}, 2018.

\bibitem{Zhang_2019_ICCV}
Xiong Zhang, Qiang Li, Hong Mo, Wenbo Zhang, and Wen Zheng.
\newblock End-to-end hand mesh recovery from a monocular rgb image.
\newblock In {\em International Conference on Computer Vision}, 2019.

\bibitem{zhou2020monocular}
Yuxiao Zhou, Marc Habermann, Weipeng Xu, Ikhsanul Habibie, Christian Theobalt,
  and Feng Xu.
\newblock Monocular real-time hand shape and motion capture using multi-modal
  data.
\newblock In {\em Conference on Computer Vision and Pattern Recognition}, 2020.

\bibitem{zimmermann2017learning}
Christian Zimmermann and Thomas Brox.
\newblock Learning to estimate 3d hand pose from single rgb images.
\newblock In {\em International Conference on Computer Vision}, 2017.

\bibitem{zimmermann2019freihand}
Christian Zimmermann, Duygu Ceylan, Jimei Yang, Bryan Russell, Max Argus, and
  Thomas Brox.
\newblock Freihand: A dataset for markerless capture of hand pose and shape
  from single rgb images.
\newblock In {\em International Conference on Computer Vision}, 2019.

\end{thebibliography}
}

\clearpage
\appendix
\noindent \textbf{\Large{Appendix}}
\\

This is the appendix of the main text. We first provide detailed information about the proposed statistical regularization terms (Section~\ref{sec:regu_appendix}), implementation and architecture (Section~\ref{sec:imple_appendix}). Then we compare the results in camera coordinates with a fully-supervised setting in Section~\ref{sec:com_fsl}. Finally, more visualization results are presented in Section ~\ref{sec:quali}.
\section{Statistical Regularization}
\label{sec:regu_appendix}
We introduce three regularization terms, including the texture regularization, the scale regularization, and the skeleton regularization (Section~3.4.1 in the main text), to make the output 3D hands more reasonable. The details are as follows.

\noindent \textbf{Texture Regularization.}\quad
Since the skin color of hands typically is uniform, we propose a texture regularization term $E_{C}$ to penalize outlier RGB values, where $f^C(c)$ is used to compute per-vertex color loss:
\begin{equation}
    E_{C}=\frac{{con}_{sum}}{n}\sum_{i=1}^{n}{f^C(c_i)}
\end{equation}
\begin{equation}
  f^C(c) = 
  \begin{cases}
    0, &\text{if $\bar{c}-2\sigma_{c}<c<\bar{c}+2\sigma_{c}$,}\\
	{\parallel c-\bar{c} \parallel}^2_2, &\text{else,}
  \end{cases}
\end{equation}
Here, $\bar{c}\in \mathbb{R}^{3}$ is the average RGB of all vertices and $\sigma_{c}\in \mathbb{R}^{3}$ is the standard deviation for three color channels.

\noindent \textbf{Scale Regularization.}\quad
The scale regularization term is used to constrain the length of the hand bone to provide a reference for the scale uncertainty in this monocular 3D reconstruction task. For each dataset, we define an average bone length $\bar{l}$ of the proximal phalanx of the middle finger (the bone $\overrightarrow{9\underline{10}}$ that between the 9th joint and the 10th joint in Fig.~3A of the main text). The scale regularization term is defined as $E_{s}=\parallel l-\bar{l} \parallel^2$ to encourage the length $l$ of the estimated hand model's proximal phalanx of the middle finger to be close to the average length $\bar{l} \in \mathbb{R}^{1}$. We 
empirically set $\bar{l}=2.82\textit{cm}$.

\begin{figure}
    \centering
    \includegraphics[width=1.05\linewidth]{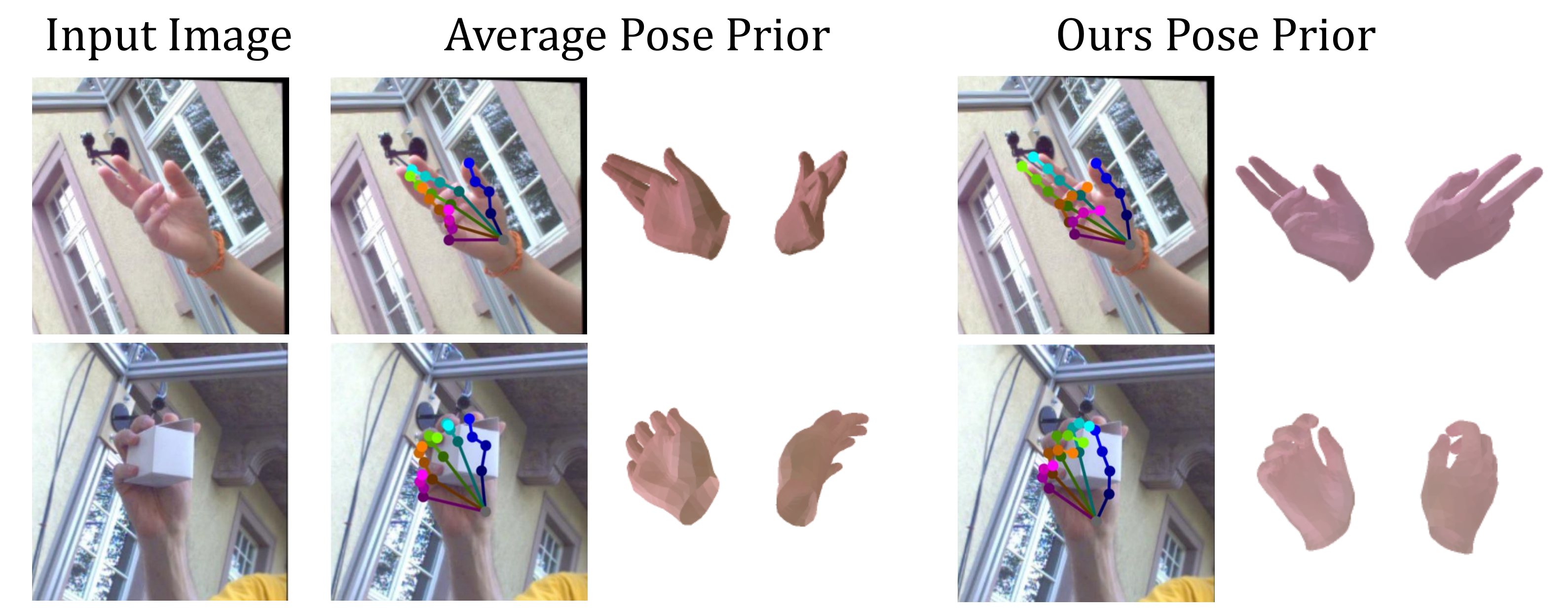}
    \caption{Qualitative comparison of the average pose prior \cite{hasson2019learning} and ours pose prior. We give two samples with the input image, projected output keypoints, and 3D mesh in two viewpoints.}
    \label{fig:posefailure}
\end{figure}

\noindent \textbf{Skeleton Regularization.}\quad
The skeleton regularization term is used to penalize invalid hand pose. Instead of using a regularization on pose parameters $\theta$ to make the pose to be close to the average pose \cite{hasson2019learning} (we call it the average pose prior), we think that feasible poses at different distances from the average pose should be treated equally without any penalty. We therefore define the feasible range $[{min}_i,{max}_i]$ (Table~\ref{tab:minmax}) for each rotation angle $a_i$ (as shown in Fig.~3B of the main text) and then penalize those who exceed the threshold:
\begin{equation}
    E_{J}=\frac{1}{15}\sum_{i=1}^{15}{f^J_i(a_i)}
\end{equation}
\begin{equation}
  f^J_i(a) = 
  \begin{cases}
	{min}_i-a, &\text{if $a \le {min}_{i},$}\\
	0, &\text{if ${min}_{i}\le a \le {max}_{i}$,}\\
	a-{max}_i, &\text{if $a \ge {max}_{i},$}
  \end{cases}
\end{equation}
 As shown in Fig.~\ref{fig:posefailure}, we give some samples of using the average pose prior and ours pose prior. When using the average pose prior, the projected output joints may be reasonable but the hand configures in 3D shape is not valid. We think this is because the average pose prior penalizes all poses according to their distance from the average pose, which cannot distinguish valid or invalid hand configure. While we only penalize the invalid hand pose and also determine this penalty term according to the degree of deviation, which results in much better performance.

\begin{table}[]
    \centering
    \begin{tabular}{c|ccc}
         \Xhline{1pt}
         \rowcolor[rgb]{ .873,  .91,  0.95}
         Bone & Azimuth & Pitch & Roll\\
         \hline
         $\overrightarrow{1\underline{2}}$ & (-22.5,33.75) & (-22.5,22.5) & (0,90)\\
         \rowcolor[rgb]{ .94,  .94,  .94}
         $\overrightarrow{2\underline{3}}$ & (-5,5) & (-22.5,22.5) & (-5,5)\\
         $\overrightarrow{3\underline{4}}$ & (-5,5) & (-100,20) & (-5,5)\\
         \rowcolor[rgb]{ .94,  .94,  .94}
         $\overrightarrow{5\underline{6}}$ & (-10,10) & (-100,10) & (-5,5)\\
         $\overrightarrow{6\underline{7}}$ & (-5,5) & (-100,10) & (-5,5)\\
         \rowcolor[rgb]{ .94,  .94,  .94}
         $\overrightarrow{7\underline{8}}$ & (-5,5) & (-100,10) & (-5,5)\\
         $\overrightarrow{9\underline{10}}$ & (-10,10) & (-100,10) & (-5,5)\\
         \rowcolor[rgb]{ .94,  .94,  .94}
         $\overrightarrow{10\underline{11}}$ & (-5,5) & (-100,10) & (-5,5)\\
         $\overrightarrow{11\underline{12}}$ & (-5,5) & (-100,10) & (-5,5)\\
         \rowcolor[rgb]{ .94,  .94,  .94}
         $\overrightarrow{13\underline{14}}$ & (-10,10) & (-100,10) & (-5,5)\\
         $\overrightarrow{14\underline{15}}$ & (-5,5) & (-100,10) & (-5,5)\\
         \rowcolor[rgb]{ .94,  .94,  .94}
         $\overrightarrow{15\underline{16}}$ & (-5,5) & (-100,10) & (-5,5)\\
         $\overrightarrow{17\underline{18}}$ & (-10,20) & (-100,10) & (-20,5)\\
         \rowcolor[rgb]{ .94,  .94,  .94}
         $\overrightarrow{18\underline{19}}$ & (-5,5) & (-100,10) & (-5,5)\\
         $\overrightarrow{19\underline{20}}$ & (-5,5) & (-100,10) & (-5,5)\\
         \Xhline{1pt}
    \end{tabular}
    \vspace{-0.2cm}
    \caption{The minimum and maximum values in \textit{degrees} of the joint angle parameters used in our proposed skeleton regularization term.}
    \label{tab:minmax}
\end{table}


\section{Implementation and Architecture Details}
\label{sec:imple_appendix}
For the weighting factors in Section~3.4 of the main text, we set \mbox{$w_{3d}=1$}, $w_{2d}=0.001$, $w_{con}=0.0002$, $w_{geo}=0.001$, $w_{photo}=0.005$, $w_{regu}=0.01$, $w_{ori}=100$, $w_{SSIM}=0.2$, $w_{C}=0.5$, $w_{s}=10000$, and $w_{J}=10$. For the HO-3D, we use the 2D keypoints information to crop the hand region from the raw image and then resize the cropped image into $224\times224$ as training samples and rely on the provided 2D bounding box to crop the testing frames. We don't apply any data augmentation for the FreiHAND.

The 3D reconstruction network has an encoder-decoder architecture. The EfficientNet-b0 \cite{tan2019efficientnet} encodes the input image $I \in \mathbb{R}^{224\times224\times3}$ to a latent feature map ${m}_{h} \in \mathbb{R}^{7\times7\times1536}$, where we also take an intermediate feature map ${m}_{l} \in \mathbb{R}^{56\times56\times32}$. A vector $v_h\in \mathbb{R}^{1536}$ is got from ${m}_{h}$ through max pooling and then passed through a series of fully connected layers $f_{base}$. Then multiple heads ($f_{pose},f_{shape},f_{trans},f_{rot},f_{scale}$) are used to estimate pose $\theta$, shape $\beta$, translation $T$, rotation $R$ and scale $s$ (Table~\ref{tab:handregressor}). We use a series of 2D convolution layers $f_{conv}$ and two heads $f_{tex},f_{light}$ to encode the higher resolution feature ${m}_{l}$ into the hand texture $C$ and scene lighting $L$ (Table~\ref{tab:tex_light}).

\begin{table}[]
    \centering
    \begin{tabular}{c|cc}
        \Xhline{1pt}
        Stage & Operator & Output \\
        \hline
        \multirow{2}{*}{$f_{base}$} & Linear(1536,1024),BN,ReLU & $1\times1024$\\
        & Linear(1024,512),BN,ReLU & $1\times512$\\
        \hline
        \multirow{2}{*}{$f_{pose}$} & Linear(512,128),ReLU & $1\times128$\\
        & Linear(128,30) & $1\times30$\\
        \hline
        \multirow{2}{*}{$f_{shape}$} & Linear(512,128),ReLU & $1\times128$\\
        & Linear(128,10) & $1\times10$\\
        \hline
        \multirow{3}{*}{$f_{trans}$} & Linear(512,128),ReLU & $1\times128$\\
        & Linear(128,32) & $1\times32$\\
        & Linear(128,3) & $1\times3$\\
        \hline
        \multirow{3}{*}{$f_{rot}$} & Linear(512,128),ReLU & $1\times128$\\
        & Linear(128,32) & $1\times32$\\
        & Linear(128,3) & $1\times3$\\
        \hline
        \multirow{3}{*}{$f_{scale}$} & Linear(512,128),ReLU & $1\times128$\\
        & Linear(128,32) & $1\times32$\\
        & Linear(128,1) & $1\times1$\\
        \Xhline{1pt}
    \end{tabular}
    \vspace{-0.2cm}
    \caption{The hand regressor architecture. Linear transformation layers are given as Linear (in\_size, out\_size).}
    \label{tab:handregressor}
\end{table}

\begin{table}[]
    \centering
    \scalebox{1}{
    \begin{tabular}{c|cc}
        \Xhline{1pt}
        Stage & Operator & Output \\
        \hline
        \multirow{5}{*}{$f_{conv}$} & Conv2d(32,48,10,4,1), ReLU & $13\times13\times48$\\
        & MaxPool(3,2) & $6\times6\times48$\\
        & Conv2d(48,64,3,1,0), ReLU & $4\times4\times64$\\
        & MaxPool(2,2) & $2\times2\times64$\\
        & Flatten & $1\times256$\\
        \hline
        \multirow{2}{*}{$f_{tex}$} & Linear(256,64),ReLU & $1\times64$\\
        & Linear(64,2334) & $1\times2334$\\
        \hline
        \multirow{2}{*}{$f_{light}$} & Linear(256,64),ReLU & $1\times64$\\
        & Linear(64,11) & $1\times11$\\
        \Xhline{1pt}
    \end{tabular}
    }
    \vspace{-0.2cm}
    \caption{The texture and lighting regressor architecture. Convolution parameters are given as Conv2d (in\_channels, out\_channels, kernel\_size, stride, padding). The 2D max pooling is given as MaxPool (kernel\_size, stride). The linear transformation layer is given as Linear (in\_size, out\_size).}
    \label{tab:tex_light}
\end{table}

\begin{table}[h]
    \centering
    \scalebox{0.8}{
    \begin{tabular}{c|cc|ccccc}
        \Xhline{1pt}
        \multirow{2}{*}{\diagbox{Supervision}{Method}} & \multicolumn{2}{c|}{MANO-CNN} & \multicolumn{2}{c}{Ours} \\
        \cline{2-5}
        & MPJP$\downarrow$ & MPVPE$\downarrow$  & MPJPE$\downarrow$ & MPVPE$\downarrow$\\
        \hline
        FSL & 8.72 & 8.84 & 8.66 & 8.77\\
        SSL & 12.75 & 12.81 & 10.57 & 10.60\\ 
        \Xhline{1pt}
    \end{tabular}
    }
    \vspace{-0.2cm}
    \caption{Comparison of unaligned results of MANO-CNN and \textit{Ours} under fully-supervision (FSL) or self-supervision (SSL) on the FreiHAND testing set.
    }
    \label{tab:fslssl}
\end{table}

\begin{table*}[thb]
\centering
\begin{tabular}{c|c|cc|cc}
\Xhline{1pt}
\rowcolor[HTML]{ECF4FF} 
\cellcolor[HTML]{ECF4FF}                   & \cellcolor[HTML]{ECF4FF}                           & \multicolumn{2}{c|}{\cellcolor[HTML]{ECF4FF}Predicted} & \multicolumn{2}{c}{\cellcolor[HTML]{ECF4FF}Projected}                  \\
\rowcolor[HTML]{ECF4FF} 
\multirow{-2}{*}{\cellcolor[HTML]{ECF4FF}\diagbox{Evaluation Matrix}{Keypoint Set}} & \multirow{-2}{*}{\cellcolor[HTML]{ECF4FF}OpenPose} & w/o 2D-3D               & w/ 2D-3D                    & \multicolumn{1}{l}{\cellcolor[HTML]{ECF4FF}w/o 2D-3D} & w/ 2D-3D       \\
\hline
\rowcolor[HTML]{EFEFEF} 
Per Joint                                  & 0.807                                              & 0.820                   & \textbf{0.828}              & 0.808                                                 & \underline{0.825}    \\
Per Frame Mean                             & 0.799                                              & 0.815                   & \textbf{0.825}              & 0.805                                                 & \underline{0.823}    \\
\rowcolor[HTML]{EFEFEF} 
Per Frame Max                              & 0.466                                              & 0.517                   & \underline{0.545}                 & 0.543                                                 & \textbf{0.577}\\
\Xhline{1pt}
\end{tabular}
\caption{Comparison of the AUC (higher is better) of 2D keypoint sets used or outputted at the training stage on FreiHAND in different evaluation matrix, where the 2D error value is from 0 to 50 pixel.}
\label{tab:2d-comps}
\end{table*}

\section{Comparison to using GT 3D as Supervision}
\label{sec:com_fsl}
In all of our experiments, we do not use the scale information (provided by the FreiHAND testing set) or the depth information (provided by the HO-3D testing set). It is typical to evaluate the 3D hand pose and shape estimation in the hand-centric coordinate (e.g., using Procrustes alignment), but it is also important to accurately output 3D hands with accurate position and scale. To this end, we show our results in camera coordinates of the proposed self-supervised (SSL) method and the results of a fully-supervised (FSL) scheme. Note that this section reports the results of raw output without using Procrustes alignment. 
For FSL, we use ground truth 2D keypoint annotations whose keypoint confidences are set to be the same, and a 3D joint loss is additionally used to give real 3D supervision. The 3D joint loss enforces the $k=21$ of output joints $J=\lbrace j_{i}\in \mathbb{R}^{3} \vert 1\le i \le k\rbrace$ and ground truth joints $J^{gt}=\lbrace j^{gt}_{i}\in \mathbb{R}^{3} \vert 1\le i \le k\rbrace$ to be aligned. We add the $E_{3dj}$ to $E$ by a weighting factor $w_{3dj}=100$.
\begin{equation}
    E_{3dj} = \frac{1}{k}\sum_{i=1}^{k}{\parallel j_{i}- j^{gt}_{i} \parallel}^2_2
\end{equation}

A big advantage of using 3D annotation is that it can help the network to learn the depth value of the output hand. In Table~\ref{tab:fslssl}, we compare unaligned results of our method with MANO-CNN under the FSL and SSL settings. Both approaches get much better performance under FSL than SSL, and our approach outperforms MANO-CNN under FSL and SSL (a 0.06\textit{cm} decrease in MPJPE under FSL and a 2.18\textit{cm} decrease in MPJPE under SSL). We further find that when degrading the supervision strength from FSL to SSL, our method shows less performance degradation than MANO-CNN, where the MPJPE of our method increases by 22.1\% while by 46.2\% in MANO-CNN. We think this is because our approach is robust to the self-supervised setting than MANO-CNN.

\section{More Comparison of Different 2D Keypoint Sets}
\begin{figure}
    \centering
    \includegraphics[width=1\linewidth]{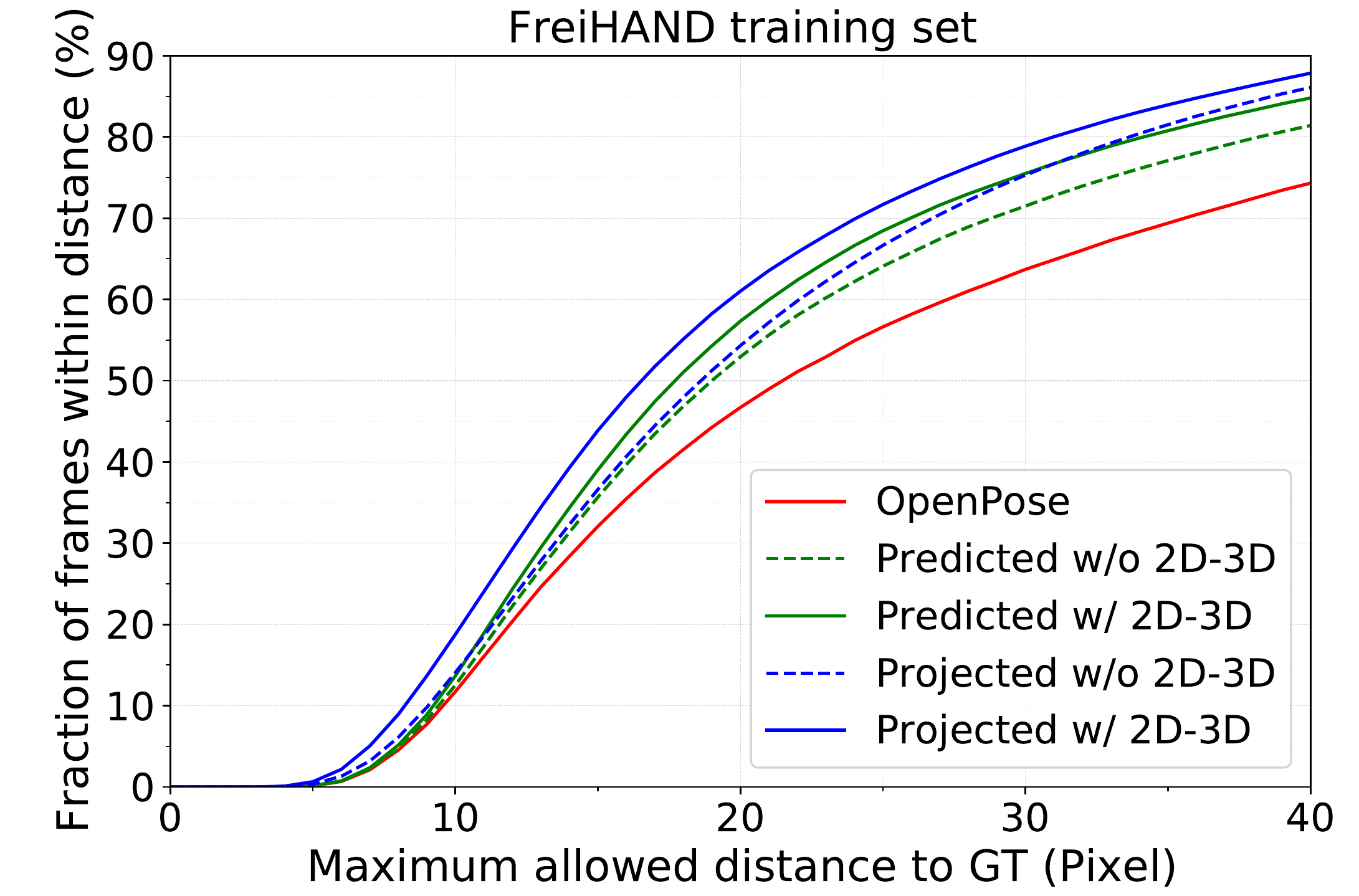}
    \caption{A comparison of 2D keypoint sets used or outputted at the training stage on FreiHAND. The fraction of frames within maximum joint distance is plotted.}
    \label{fig:max-2d}
\end{figure}

In Section 4.4.2 and Fig 4 of the main body, we compared the fraction of ``Per Joint" 2D error. 
Here, we further visualize the fraction of ``Per Frame Max" 2D error, i.e., the fraction of frames within maximum 2D MPJPE in \textit{pixel}, in Fig.~\ref{fig:max-2d}. The AUC (0$\sim$50 pixel) of ``Per Joint" error, ``Per Frame Mean" error and ``Per Frame Max" error in 2D are shown in Table~\ref{tab:2d-comps}.

As shown in Fig.~\ref{fig:max-2d} and Table~\ref{tab:2d-comps}, the 2D-3D consistency loss $E_{con}$ improves the AUC of both the 2D branch (from \textbf{\textit{Predicted w/o 2D-3D}} to \textbf{\textit{Predicted w/ 2D-3D}}) and the 3D branch (from \textbf{\textit{Projected w/o 2D-3D}} to \textbf{\textit{Projected w/ 2D-3D}}).
All results of the 2D and 3D branches outperform \textbf{\textit{OpenPose}} which is used as the keypoint supervision source.

As shown in Table~\ref{tab:2d-comps}, in term of ``Per Joint" and ``Per Frame Mean", the predicted results from the 2D branch are better than the projected results from the 3D branch since the 2D branch is designed for 2D keypoints estimation while the results of the 3D branch are projected from 3D outputs. 
While in terms of ``Per Frame Max", we find that the projected results are better than the predicted ones. 
We believe this is due to the constraints contained in the MANO model, the projected results eliminate outlier 2D results.
Thus, the AUC of ``Per Frame Max", which is more sensitive to outliers, is greatly improved from OpenPose to Predicted \& Projected due to $E_{con}$ and the powerful regularization characteristic of network training (Section~4.4.2 of the main body).


\section{Qualitative Results}
\label{sec:quali}
\begin{figure*}
    \centering
    \includegraphics[width=1\linewidth]{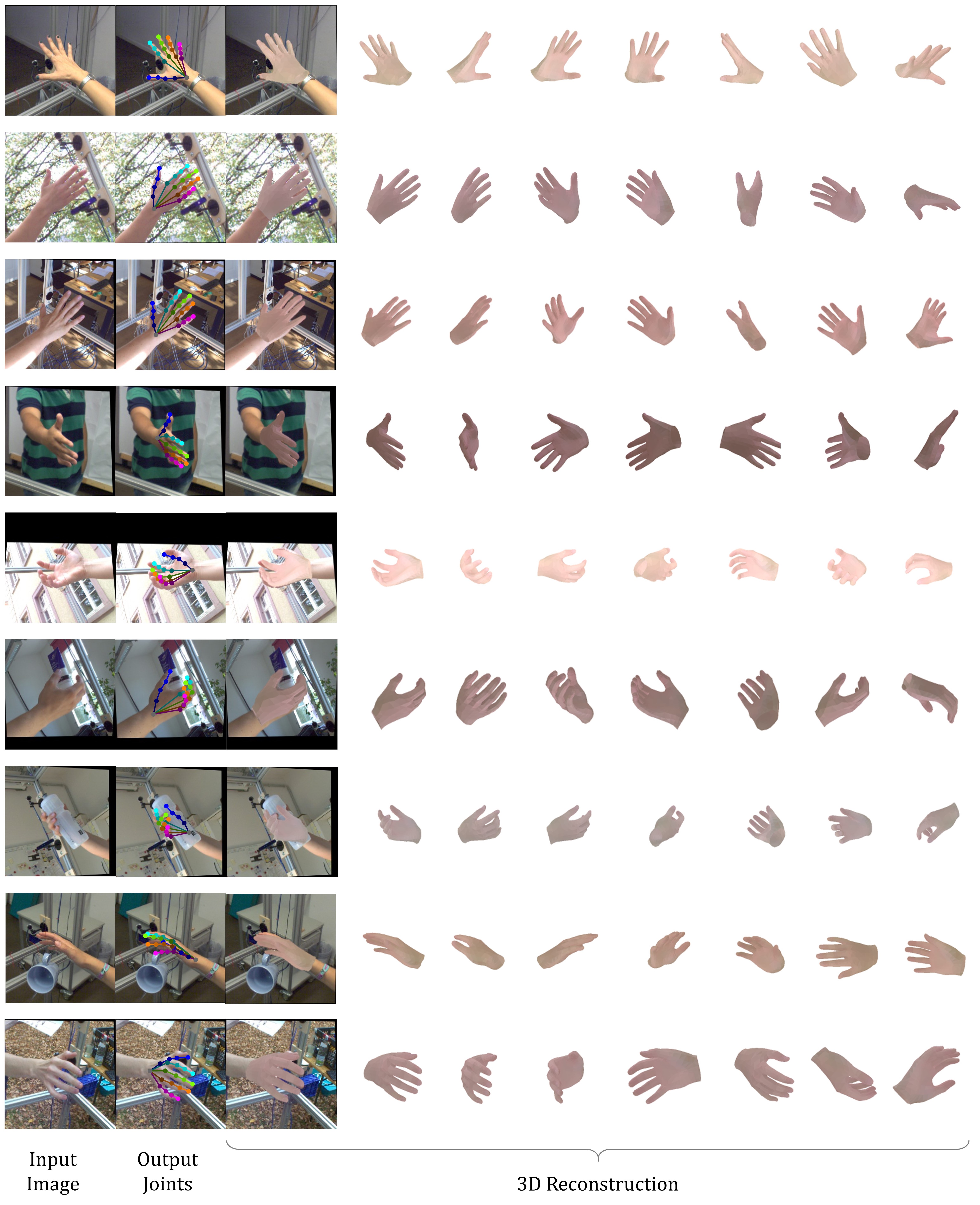}
    \caption{Qualitative visualization of our method on the FreiHAND testing set (Part 1). From left to right: input image, output 3D joints projected to image space, output 3D mesh projected to image space, 3D mesh from different views (4th-10th column).}
    \label{fig:freihand0}
\end{figure*}

\begin{figure*}
    \centering
    \includegraphics[width=1\linewidth]{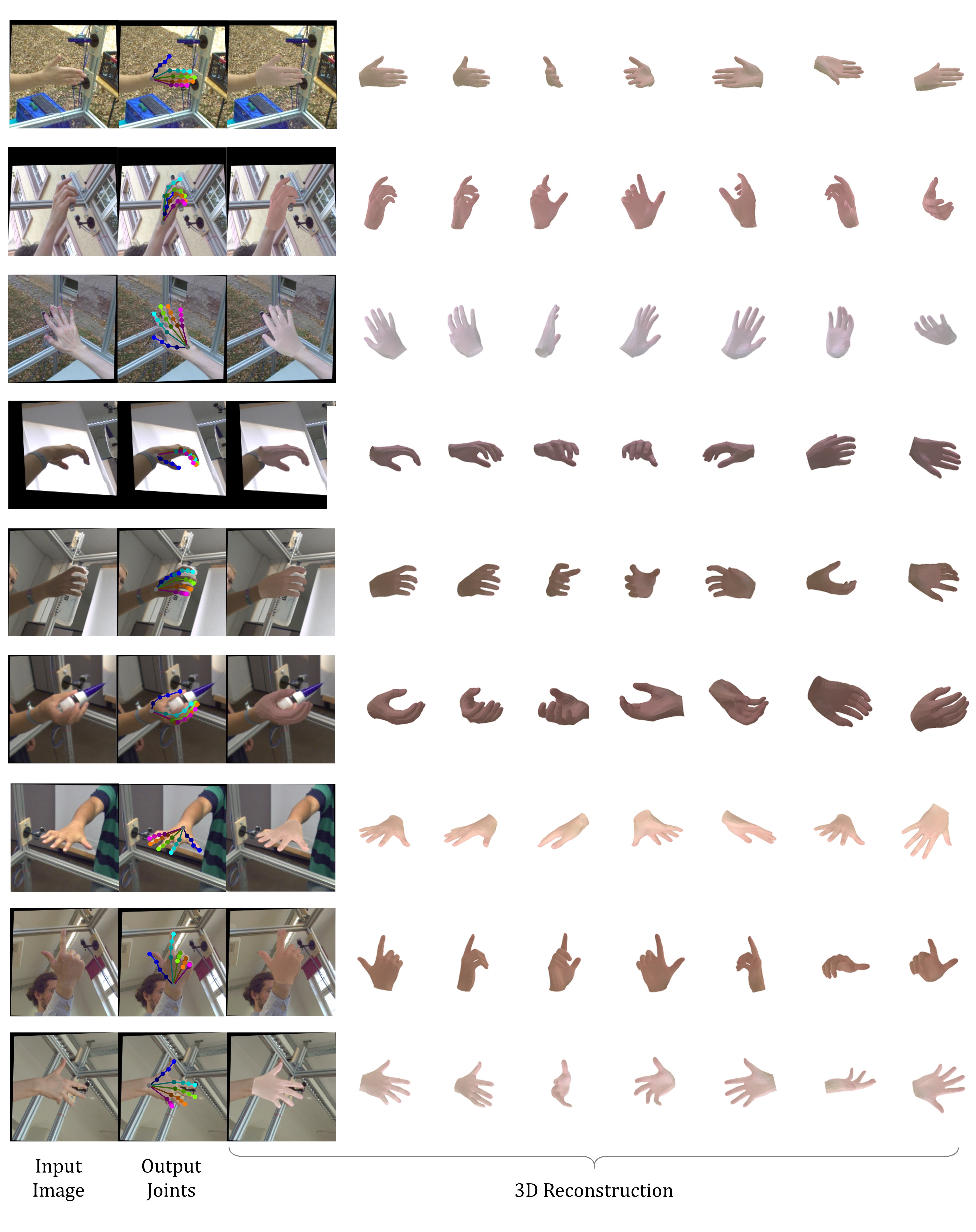}
    \caption{Qualitative visualization of our method on the FreiHAND testing set (Part 2). From left to right: input image, output 3D joints projected to image space, output 3D mesh projected to image space, 3D mesh from different views (4th-10th column).}
    \label{fig:freihand1}
\end{figure*}

\begin{figure*}
    \centering
    \includegraphics[width=1\linewidth]{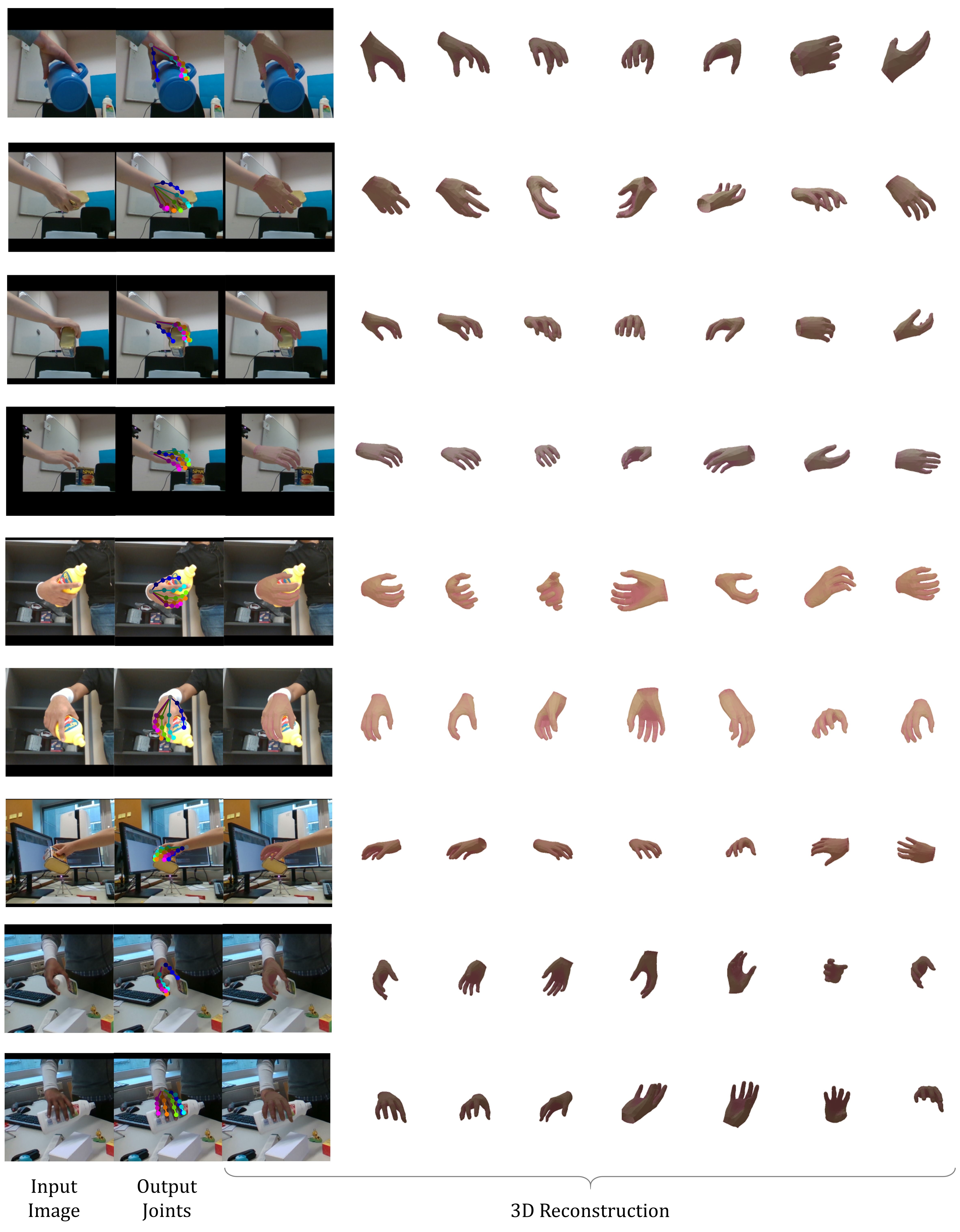}
    \caption{Qualitative visualization of our method on the HO-3D testing set. From left to right: input image, output 3D joints projected to image space, output 3D mesh projected to image space, 3D mesh from different views (4th-10th column).}
    \label{fig:ho3d}
\end{figure*}

We provide more qualitative results of single-view 3D hand reconstruction. Fig.~\ref{fig:freihand0}~and~\ref{fig:freihand1} report the qualitative results of samples from the FreiHAND testing set. Fig.~\ref{fig:ho3d} shows the qualitative results of samples from the HO-3D testing set. Note that hands in HO-3D suffer from more serious occlusion, resulting in more object or background pixels in the masked foreground during the self-supervised texture learning. 
So the texture estimation is less accurate on the HO-3D.

\begin{figure*}
    \centering
    \vspace{-0.5cm}
    \includegraphics[width=0.85\linewidth]{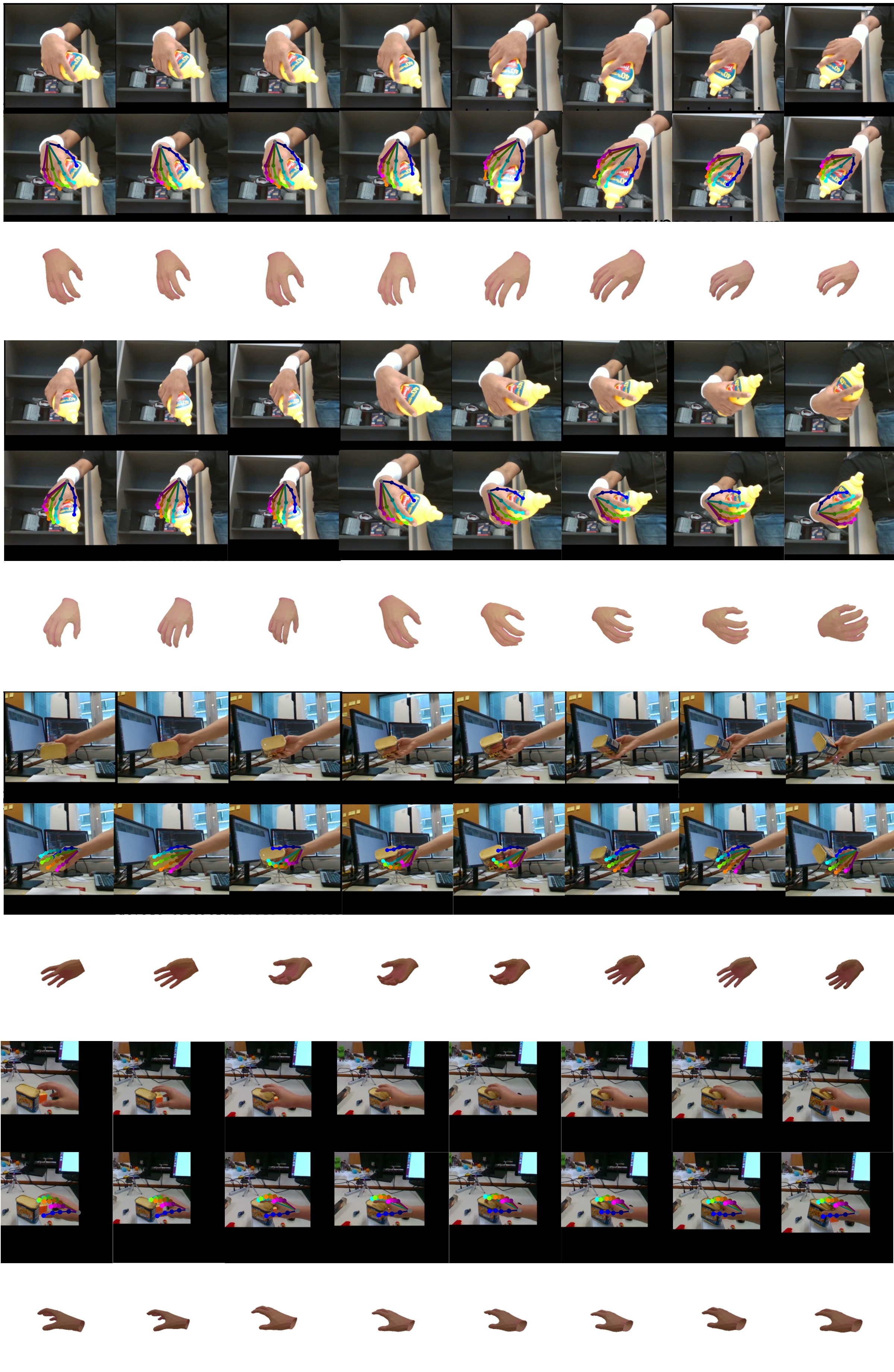}
    \caption{Qualitative visualization of our method on the HO-3D testing set. We give hand reconstruction results for four sequences. Eight samples are shown in each sequence, and we visualize the input image, projected output joints, and output mesh for each sample.}
    \label{fig:seq-results}
\end{figure*}

Since the HO-3D provides video sequences in the testing set, we visualize the results of sequence frames on the HO-3D testing set in Fig.~\ref{fig:seq-results}.
Although we do not use temporal information during the training and only use a single image for inference, our model outputs accurate shape with consistent shape and texture for each video sequence.

\end{document}